%% file: main.tex
\documentclass[10pt,twocolumn,letterpaper]{article}

\usepackage[pagenumbers]{iccv} %

\input{preamble}
\usepackage{pifont}
\newcommand{\cmark}{\ding{51}}%
\usepackage{gradient-text}
\newcommand{\bigo}[1]{\ensuremath{\mathcal{O}(#1)}}
\usepackage[table,xcdraw,dvipsnames]{xcolor} %
\newcommand{\gr}{\rowcolor[gray]{.92}} %
\newcommand{\grc}[1]{\cellcolor[gray]{.92}#1} %
\usepackage{tabularx}
\usepackage{bm}
\usepackage{tikz}
\usepackage{multirow}
\usepackage{balance}
\definecolor{iccvblue}{rgb}{0.21,0.49,0.74}
\usepackage[pagebackref,breaklinks,colorlinks,allcolors=iccvblue]{hyperref}

\def\paperID{10288} %
\def\confName{ICCV}
\def\confYear{2025}

\newcommand{\sotacolor}[1]{\textbf{\gradientRGB{#1}{244,33,193}{54,67,153}}}

\newcommand{\spa}[1]{\textbf{\gradientRGB{#1}{244,33,193}{120,50,173}}}
\newcommand{\colorR}[1]{\textbf{\gradientRGB{#1}{120,50,173}{87,58,163}}}
\newcommand{\colorC}[1]{\textbf{\gradientRGB{#1}{87,58,163}{54,67,153}}}

\newcommand{\name}{SpaRC} %

\title{\sotacolor{SpaRC}: \spa{Spa}rse \colorR{R}adar-\colorC{C}amera Fusion for 3D Object Detection}

\author{Philipp Wolters\quad %
Johannes Gilg\quad
Torben Teepe\quad
Fabian Herzog\quad
Felix Fent\quad
Gerhard Rigoll\quad
\\ [0.15cm]
Technical University of Munich \qquad
}

\begin{document}
\maketitle
\input{sec/0_abstract}
\input{sec/1_intro}

\input{sec/2_related_work}

\input{sec/3_method}

\input{sec/4_experiments}

\input{sec/6_conclusion}

{
    \small
    
    \bibliographystyle{ieeenat_fullname}
    \bibliography{main}
}
\input{sec/7_suppl}

\end{document}

%% file: sec/0_abstract.tex
\begin{abstract}

     In this work, we present \textbf{SpaRC}, a novel \textbf{Spa}rse fusion transformer for 3D perception that integrates multi-view image semantics with \textbf{R}adar and \textbf{C}amera point features.
    The fusion of radar and camera modalities has emerged as an efficient perception paradigm for autonomous driving systems.
    While conventional approaches utilize dense Bird's Eye View (BEV)-based architectures for depth estimation, contemporary query-based transformers excel in camera-only detection through object-centric methodology.
    However, these query-based approaches exhibit limitations in false positive detections and localization precision due to implicit depth modeling.
    We address these challenges through three key contributions: (1) sparse frustum fusion (SFF) for cross-modal feature alignment, (2) range-adaptive radar aggregation (RAR) for precise object localization, and (3) local self-attention (LSA) for focused query aggregation.
    In contrast to existing methods requiring computationally intensive BEV-grid rendering, SpaRC operates directly on encoded point features, yielding substantial improvements in efficiency and accuracy.
    Empirical evaluations on the nuScenes and TruckScenes benchmarks demonstrate that SpaRC significantly outperforms existing dense BEV-based and sparse query-based detectors.
    Our method achieves state-of-the-art performance metrics of \textbf{67.1 NDS} and \textbf{63.1 AMOTA}.
    The code and pretrained models are available at \url{https://github.com/phi-wol/sparc}.
\end{abstract}

%% file: sec/1_intro.tex
\section{Introduction}
\label{sec:intro}

Developing an efficient, robust, and scalable perception system for autonomous driving is a challenging task. 
Autonomous vehicles must accurately perceive their surroundings and make informed decisions in real-time to ensure safe operation in complex dynamic environments like crowded urban scenarios and fast-paced highways.
This requires precise localization and classification of other traffic participants \cite{hu2023planning, li2023delving}.
\begin{figure}[t]
    \centering
    \includegraphics[width=\linewidth]{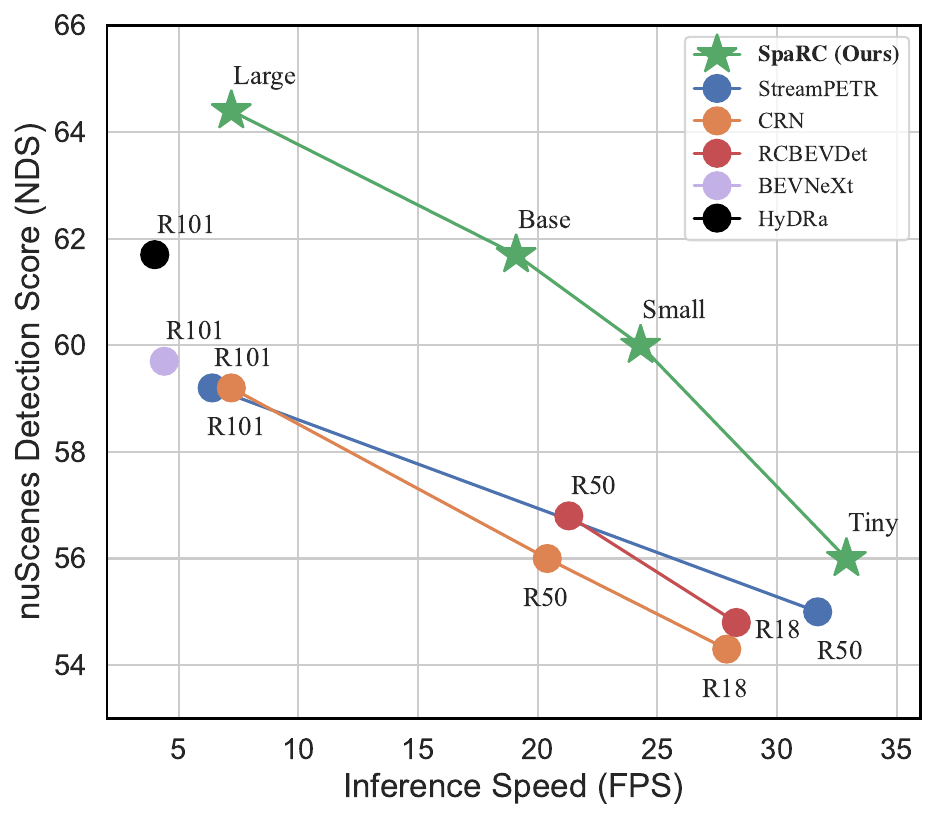}
    \caption{Compared to previous real-time perception models our \textbf{\sotacolor{\name}}~model family achieves state-of-the-art performance in accuracy and inference speed. The inference speed is measured with a single consumer-grade RTX3090 GPU on nuScenes val.}
    \label{fig:inference_analysis}
\end{figure}
Multi-modal sensor fusion of LiDAR, camera, and radar has made significant progress in recent years due to large-scale, diverse datasets \cite{geiger2012we, caesar2020nuscenes, sun2020waymo, wilson2023argoverse} and advances in deep learning architectures \cite{yin2021center, li2023delving, liang2022bevfusion,liu2022petr}.
While LiDAR-based methods achieve impressive performance \cite{liang2022bevfusion, chen2023focalformer3d, wang2023dsvt}, their high cost and maintenance requirements limit widespread deployment. 
This has motivated research into more cost-effective sensor combinations, particularly camera-radar fusion \cite{bang2024radardistill,kim2023crn,wolters2024hydra,lin2024rcbevdet}.

Cameras provide high-resolution semantic information and capture rich texture details but struggle with depth estimation and perform poorly in adverse lighting and weather conditions like night, fog, or snow \cite{yoneda2019automated}. 
In contrast, Millimeter-wave radar sensors offer sparse, metric range sensing and Doppler-based velocity measurements even under adverse lighting and weather conditions. 
Combined in a complementary architecture, they have the potential to unlock reliable and affordable 3D perception for autonomous driving \cite{zhou2023bridging}.
The main barrier to radar-based perception has been the lack of high-quality and large-scale sensor recordings. 
Out of the traditional autonomous driving datasets~\cite{geiger2012we, caesar2020nuscenes, sun2020waymo, wilson2023argoverse}, only nuScenes offers a limited and noisy radar sensor suite. %
With new datasets like TruckScenes \cite{fent2024man} and the BSD dataset \cite{armanious2024bosch} this is going to change.
They offer modern 4D imaging radar sensors, long-range annotations up to 150m, and a diverse set of scenarios, including various weather conditions.
However, bridging the view disparity between the dense camera images and the sparse radar representation remains a key challenge due to the unique characteristics of the radar sensor \cite{zhou2023bridging}:
low angular resolution, only a few reflected points per object, noise, and clutter due to multi-path reflections capture the intricacies of radar-based perception and require an adaptive fusion design \cite{fan20244d}.

Most existing methods are based on LiDAR-centric architectures that utilize dense point-cloud processing backbones like PointPillars \cite{lang2019pointpillars} with Bird's-Eye-View (BEV) feature extraction and fusion mechanisms, which have become the default choice for 3D object detection \cite{liu2022bevfusion, liang2022bevfusion, yan2023cmt, yang2022deepinteraction,huang2023detecting}. 
However, directly applying these dense BEV representations to sparse radar data leads to computational inefficiency, as most grid cells remain empty.
Recent work has focused on adapting these LiDAR-centric designs for camera-radar fusion through various strategies: radar-aided depth estimation \cite{long2023radiant,kim2023crn,wolters2024hydra}, modified grid-rendering backbones \cite{lin2024rcbevdet,musiat2024radarpillars}, and adaptive BEV fusion mechanisms \cite{kim2023crn,lin2024rcbevdet,wolters2024hydra}. 
Even with recent improvements in BEVFusion-style architectures, this fundamental mismatch between dense representations and sparse radar signals remains a key limitation for efficient optimization.

In contrast, we propose a novel query-based fusion transformer for 3D object detection that concentrates computational resources on salient regions of the radar modality.
We disregard the BEV-grid representation due to its sparseness in feature representation and opt for an object-centric paradigm.
Introducing \textbf{\name}, we achieve a new state-of-the-art in camera-radar 3D perception with strong robustness, high accuracy, and real-time inference speed.
Our main contributions are:
\begin{itemize}[leftmargin=+.5cm]
    \item We utilize a modality-specific sparse feature set representation for radar encoding. %
    \item We design a multi-scale but Sparse Frustum Fusion for efficient cross-modal feature alignment, improving the projection-based representation and explicit depth estimation.
    \item We propose a range-adaptive radar refinement and a local self-attention mechanism to model the intuitive object-to-point interactions and improve the implicit depth learning.
    \item \textbf{\name}~achieves state-of-the-art performance on the nuScenes benchmark ($+2.9$ NDS and $+2.6$ mAP). Moreover, our findings generalize to the long-range and adverse conditions on the new TruckScenes benchmark and match the LiDAR-based baseline.
\end{itemize}

%% file: sec/2_related_work.tex
\section{Related Work }
\label{sec:related_work}

\subsection{Dense BEV-based 3D Perception} %

Since the seminal work of LSS \cite{philion2020lift}, vision-centric 3D perception has moved from the perspective view \cite{wang2021fcos3d,wang2022probabilistic,yin2021center} to a unified Bird's-Eye-View (BEV) space \cite{philion2020lift, li2022bevformer,huang2021bevdet, teepe2024earlybird}.
The 3D space representation has been proven to be beneficial for unified multi-view and point-cloud fusion, as well as downstream tasks such as mapping, tracking and planning~\cite{harley2022simple,zhou2019objects,hu2023planning}.
Several differentiable lifting strategies have been proposed to transform 2D image features into the BEV~\cite{harley2022simple,teepe2024lifting}.
Most prominent, the convolution-based BEVDet Series \cite{huang2021bevdet, li2022bevdepth, li2022bevstereo, park2022time} introduce efficient forward view-transformation, explicit depth prediction, and ego-motion-based temporal modeling. 
Contrary, BEVFormer \cite{li2022bevformer, yang2023bevformerV2} queries the image features using 3D-to-2D cross-attention, modeling the inverse and implicit camera un-projection.
Recent works \cite{li2023fb, li2023fbocc, li2024bevnext} combine both directions into a unified architecture to tackle the sparseness of the feature representation within the dense BEV grid.

\begin{figure*}[htbp]
    \centering
    \includegraphics[width=0.84\textwidth]{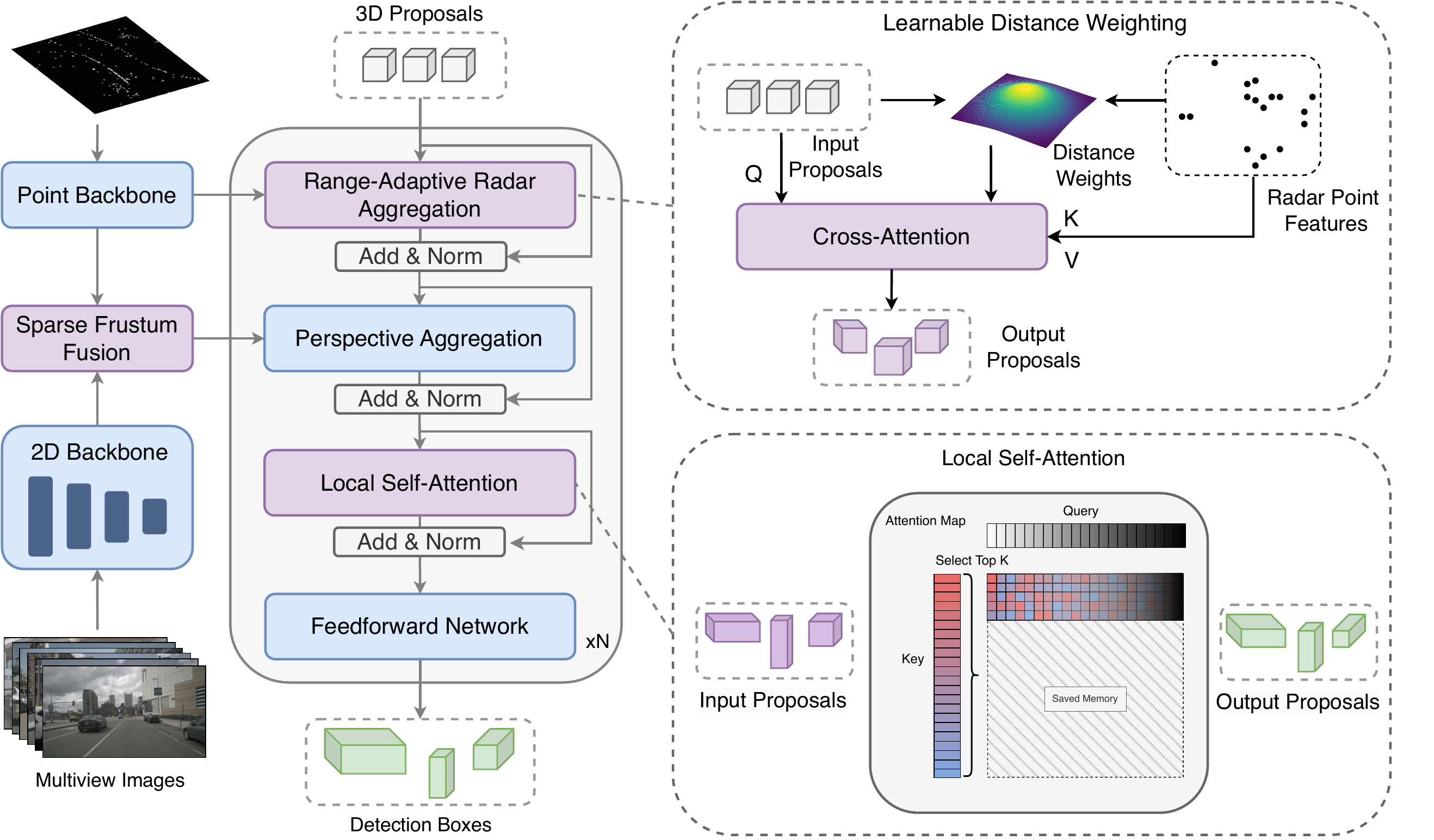}
    \caption{
        \textbf{Architecture of \name}. A fully-sparse object-centric radar-camera fusion 3D object detector.
        Radar points are sparsely encoded as a 3D-embedded feature set.
        Objects are modeled as a set of queries. 
        Range-adaptive Radar Attention encodes the queries to salient region proposals, introducing strong spatial priors and inducing object-specific Doppler velocity.
        The radar-guided deformable cross-attention is applied to frustum-fused perspective features.
        Local top K self-attention focuses the object filtering and encoding on the local region around the 3D location.
        The fused features are decoded into 3D object proposals using a sparse detection head. %
    }
    \label{fig:architecture}
\end{figure*}

\subsection{Developing Radar-Camera Fusion Systems}
Radar-camera fusion addresses the core challenge of vision-centric systems: precise and robust depth estimation.
Incorporating mmWave radar features into different stages of the detection architecture, the sparse range, and Doppler measurements reduce the overall localization errors and improve velocity estimation.
Hence, bridging the view disparity between the two feature spaces is an active area of research: 
Early works like CRF-Net \cite{nobis19crfnet}, GRIF-Net \cite{kim2020grif}, CenterFusion \cite{nabati2021centerfusion}, CRAFT \cite{kim2022craft}, and RADIANT \cite{long2023radiant} focus on projective fusion in the image space. 
These methods first project 3D radar points into the 2D image plane and then perform late-stage feature association through high-level feature concatenation or Region-of-Interest (ROI) pooling. 
The fused features are used to refine image-based object proposals by incorporating radar's precise range measurements.
More recent works have explored alternative radar feature extraction methods. 
RadarGNN \cite{fent2023radargnn} models point-pair relationships through graph neural networks, while X3KD \cite{klingner2023x3kd}, RadarDistill \cite{bang2024radardistill}, and CRKD \cite{zhao2024crkd} leverage cross-modal knowledge distillation to enhance radar feature learning.

Following the success of BEVFusion \cite{liu2022bevfusion, liang2022bevfusion, harley2022simple, man2023bev, chen2022futr3d}, dense fusion in BEV space through concatenation, summation, or SE-Blocks has emerged as the dominant paradigm. 
These methods typically pair dense BEV-based 3D object detectors with grid-based radar feature encoders like PointPillars \cite{lang2019pointpillars, li2023pillarnext}. 
The fused features are decoded into 3D object proposals using dense detection heads like CenterPoint~\cite{zhou2019objects} or TransFusion \cite{bai2022transfusion}.
Originally designed for LiDAR-centric perception, these methods have been adapted to better handle sparsity, calibration errors, and noise interference.
RCM-Fusion \cite{kim2024rcm} relies on an instance-level refinement within the dense BEV-grid.
While CRN~\cite{kim2023crn} and RCBEVDet \cite{lin2024rcbevdet} upgrade the BEV-fusion with deformable cross-attention for increased receptive fields, HyDRa \cite{wolters2024hydra} introduces a hybrid fusion, leveraging multi-modal depth estimation and a radar-guided backpropagation for refinement.

Despite their success, current BEV-based fusion methods face several key challenges. 
First, the effectiveness of BEV feature maps deteriorates significantly with distance - only about 50\% of grid cells receive valid projected image features \cite{li2023fb}. 
This sparsity is even more pronounced for radar features, where point-pillar encoders typically populate just 1-5\% of the grid cells with radar points, leading to inefficient dense representations of inherently sparse information. 
Second, state-of-the-art approaches \cite{lin2024rcbevdet,kim2023crn,wolters2024hydra} rely heavily on ego-motion-based temporal feature warping and require large receptive fields to compensate for the sparse nature of the features. 
This becomes particularly problematic for long-range perception \cite{wilson2023argoverse,fent2024man}, where computational complexity increases quadratically with range.
In contrast, our method addresses the limitations through an object-centric set-to-set fusion. 
By operating directly on sparse point-based representations rather than dense grids, we maintain information density while reducing computational overhead. 
Our point-to-point interaction focuses only on the local neighborhood of object queries, enabling stable optimization and long-range perception without the quadratic scaling of grid-based methods.

\subsection{Sparse Query-based Perception}
Sparse query-based methods have been inspired by the DEtection TRansformer (DETR) \cite{carion2020end,wang2022detr3d,zhang2023dino} and emerge as a powerful and efficient alternative to grid-based methods.
The PETR-Series \cite{liu2022petr, liu2023petrv2} models a small set of object queries with a 3D position embedding and encodes them with cross-attention.
Multi-head self-attention exhibits the role of the BEV encoder \cite{liu2023sparsebev}.
SparseBEV \cite{liu2023sparsebev} and Sparse4D \cite{lin2022sparse4d} enable spatio-temporal sampling from 3D queries in 2D feature maps by projecting deformable sampling offsets onto the 2D feature maps.
Instead of stacking ego-motion compensated BEV-grids \cite{kim2023crn,lin2024rcbevdet,wolters2024hydra}, StreamPETR \cite{Wang2023streampetr} follows up with a temporal propagation module to iteratively refine the object queries from history queries.
Far3D \cite{jiang2024far3d} shows that the sparse design is also beneficial for long-range detection with strong object recall when employing a perspective 2D object head with a depth network for dynamic query initialization.
Follow-up works \cite{liu2025ray, hou2024open, tang2025simpb} concentrate on different denoising strategies and positional encodings to increase the robustness and reduce false positives of ambiguous feature sampling along projection rays.
Due to the implicit depth modeling of the 3D-located queries, these methods achieve strong recall but suffer from false positives and localization errors.
Our object-centric radar fusion addresses the limitations of implicit localization by backprojecting and sampling from salient radar points.
This reduces false correspondences between 3D and 2D space.
The temporal and spatial filtering of queries can be reduced to focused local regions.
Doppler velocity has a synergistic effect and is naturally suited for object-level motion modeling and compensation.

%% file: sec/3_method.tex
\section{SpaRC Architecture}
\label{sec:main}

We introduce \textbf{SpaRC}, a novel \textbf{Spa}rse fusion transformer for 3D perception from \textbf{R}adar and \textbf{C}amera. 
Our model processes multi-view RGB images and radar point clouds in parallel streams: images are encoded by convolutional feature extractors with FPN \cite{lin2017feature}, while radar points are processed by a transformer-based point encoder. 
The resulting features are fused in two stages: First, radar features are projected into image space and associated with semantic feature maps. 
Second, a sparse set of 3D object queries initialized from perspective proposals and spatially distributed 3D queries, aggregates multi-modal information through cross-attention. 
Range-adaptive radar refinement guides the object-radar interaction based on distance, while deformable attention in perspective space captures the fused semantic features. 
The model maintains a high recall through its implicit design while increasing precision through strong spatial cues in the query decoder. %

\subsection{Radar Point Encoder}
Inspired by \cite{zhao2021point, wu2024point}, we employ a lightweight point transformer to extract features from radar point clouds. 
The encoder transforms unstructured radar points into a sparse but information-dense representation through space-filling curves and serialized neighbor mapping. 
By grouping points into non-overlapping patches and performing within-patch attention, we efficiently model spatial relationships without constructing and processing dense grids. 

The 3D points are encoded in the same positional embedding space as the object queries, enabling direct interaction in later fusion stages. 
To prevent overfitting on the small radar point cloud while maintaining real-time performance, we adopt a reduced version of the backbone with implementation details provided in the Appendix.
\begin{figure}[tb]
    \centering
    \includegraphics[width=1.0\linewidth]{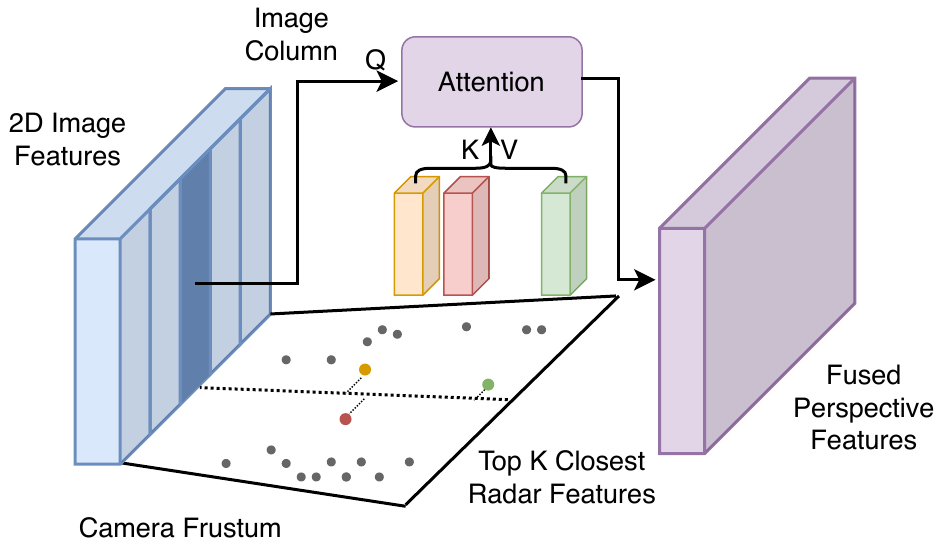}
    \caption{\textbf{Sparse Frustum Fusion Visualization.} Encoded radar points are projected into the camera frustum space and associated with semantic image feature columns. Per image column, only the K nearest points and embeddings are used for cross-attention. This enables the subsequent perspective 3D head to also benefit from the radar-fusion.}
    \label{fig:sff}
\end{figure}

\input{tables/nusc_val}
\subsection{Sparse Frustum Fusion}

As visualized in \cref{fig:sff}, we propose a Sparse Frustum Fusion (SFF) module to efficiently associate radar and image features in perspective space. 
The encoded radar feature vectors are first projected into the camera frustum space and filtered per view. 
We embed the projected coordinates (depth and horizontal pixel position) in a learnable positional encoding, while image features are embedded using their downsampled pixel positions. 
For each image column~\cite{wolters2024hydra}, we query the K nearest radar points along the vertical dimension and fuse them through cross-attention. 
This enables soft association between modalities without requiring noisy depth maps or pillar representations.
While we explicitly project radar points using their range measurements, the cross-attention mechanism allows the model to handle uncertainties from noisy measurements and missing height information.
This can be leveraged for each feature map level of the multi-scale feature extractor. 

The computational complexity of SFF is \bigo{WK}, where $W$ is the downsampled image width, and $K$ is a hyperparameter defining the number of nearest radar points per column that are considered. 
With typical values of $W,K < 100$, this leads to efficient parallel computation. 
By operating in perspective space, radar features benefit from fine-grained image supervision while preserving their 3D spatial information. 
The learnable depth and positional embeddings allow the model to explicitly align features and handle noise, establishing semantically meaningful associations between radar points and semantically connected image regions.
The subsequent perspective head can now dynamically allocate stronger 3D proposals, which initialize 3D object queries.

\subsection{Range Adaptive Radar Aggregation}

Objects are modeled as 3D reference points and semantic context features, which later are decoded into 3D bounding boxes with localization offsets, size, orientation, and velocity.
Drawing inspiration from multi-scale self-attention mechanisms \cite{liu2023sparsebev}, we propose a Range-Adaptive Radar (RAR) aggregation decoder layer that dynamically adjusts feature interactions based on spatial relationships. 
As each object is represented by a reference point in 3D space and a context embedding that encodes semantic information, we can directly associate radar points through a set-to-set interaction with the object queries and update the context embedding.

Specifically, we formulate a distance-aware attention mechanism that adaptively weights radar features based on their proximity to object centers:
\begin{equation}\label{rara}
    \text{Attn}(\mathbf{q}, \mathbf{k}, \mathbf{v}) = \text{softmax}\left(\frac{\mathbf{q}\mathbf{k}^T}{\sqrt{d}} - \alpha \frac{\|\mathbf{p}_q - \mathbf{p}_k\|_2}{r_\text{max}}\right)\mathbf{v}
\end{equation}
where query features $\mathbf{q} \in \mathbb{R}^{N_q \times d}$ interact with radar-derived key-value pairs $\mathbf{k}, \mathbf{v} \in \mathbb{R}^{N_k \times d}$ through scaled dot-product attention, modulated by a learnable distance penalty. 
Here, $\mathbf{p}_q \in \mathbb{R}^{N_q \times 3}$ and $\mathbf{p}_k \in \mathbb{R}^{N_k \times 3}$ represent the 3D positions of queries and radar points respectively, normalized by the maximum detection range $r_\text{max}$. 
The learnable parameter $\alpha$ controls the strength of the spatial bias. 

By embedding both queries and radar points in a shared continuous 3D space, RAR enables direct point-to-point interactions without requiring dense and discretized grid representations. 
This distance-guided attention mechanism naturally focuses on locally relevant radar features while maintaining the ability to capture longer-range dependencies when needed. 
The resulting radar-aware query features provide strong spatial priors that guide subsequent cross-modal fusion through deformable attention, effectively highlighting image regions that align with reliable and salient radar reflections. 

The RAR module provides two key benefits: First, it reduces local uncertainty by incorporating precise radar depth measurements before queries interact with image features
This helps resolve depth ambiguities that typically plague pure vision-based approaches.
Second, it enables object-level motion modeling by directly associating radar Doppler measurements with object queries rather than trying to infer motion from grid-based feature maps.
This sequential fusion approach - first establishing strong spatial priors through radar, then refining with dense image features - leads to more robust 3D object detection compared to methods that rely solely on back-projection or dense grid-based fusion.

\subsection{Local Self-Attention }

Traditional DETR-like architectures employ global self-attention in their decoder blocks, where each object query attends to all other queries. 
However, we observe that for 3D object detection, queries primarily need to interact with their spatial neighbors that represent the same or nearby objects. 
We propose a Local Self-Attention (LSA) mechanism that significantly improves both efficiency and effectiveness.

Our decoder processes three types of object queries: dynamically allocated queries from the perspective head, randomly initialized 3D queries, and temporal history queries from the memory queue. 
Moreover, we restructure the decoder block to apply self-attention at the end, after cross-modal feature aggregation, updating the queries from all modalities before associating them with history queries. 
This allows queries to first gather relevant features before determining their spatial relationships and filtering out the duplicates and false positives.

The key innovation of LSA is to restrict each query's attention to only its k-nearest neighbors in 3D space. For each query, we compute distances to all other queries and select the top-k closest ones as its attention context. 
This local neighborhood typically contains queries that project onto the same image regions or represent temporally consistent and propagated proposals. 
By focusing on spatially proximate queries, LSA helps establish more meaningful relationships between queries that likely correspond to the same physical object.

It significantly reduces computational complexity from \bigo{N^2} for global attention to \bigo{NK}, where $N$ is the total number of queries and $K$ is the number of neighbors. 
The distance computation is performed only once in the first decoder layer, and the same local attention pattern can be reused in subsequent layers. 
The receptive field can still grow across multiple decoder layers as information propagates through overlapping local neighborhoods. 

\subsection{Real-time Object Detection}
To achieve real-time performance, we introduce a family of models with varying complexity and speed-accuracy trade-offs. 
Our Tiny model (ResNet-18 backbone, 4 decoder layers) achieves over 30 FPS, while our Large model (ResNet-101, 6 layers) operates at 7 FPS with state-of-the-art accuracy (\cf \cref{fig:inference_analysis}). 
We employ several key optimizations:
First, we drop the perspective head from \cite{jiang2024far3d}, which significantly reduces inference time with only marginal accuracy impact. 
Second, we leverage CUDA streams to parallelize radar and camera backbone processing.
Third, our Local Self-Attention mechanism reduces computational complexity while maintaining detection quality.
Detailed architecture specifications (number of decoder layers, queries, ResNet backbones, PointTransformer settings) can be found in the Appendix.

%% file: tables/nusc_val.tex
\begin{table*}[!t]
    \centering
    \resizebox{\textwidth}{!}{
    \begin{tabular}{l|c|c|c|cc|c@{\hspace{1.0\tabcolsep}}c@{\hspace{1.0\tabcolsep}}c@{\hspace{1.0\tabcolsep}}c@{\hspace{1.0\tabcolsep}}c} 
        \toprule
        \textbf{Methods}                                    & \textbf{Input}    & \textbf{Backbone}     & \textbf{Image Size}   & \textbf{NDS$\uparrow$}    & \textbf{mAP$\uparrow$}    & \textbf{mATE$\downarrow$}     & \textbf{mASE$\downarrow$} & \textbf{mAOE$\downarrow$} & \textbf{mAVE$\downarrow$} & \textbf{mAAE$\downarrow$} \\
        
        \midrule
        CenterPoint-P \cite{yin2021center}                  & L                 & Pillars               & -                     & 59.8                      & 49.4                      & 0.320                         & 0.262                     & 0.377                     & 0.334                     & 0.198 \\
        CenterPoint-V \cite{yin2021center}                  & L                 & Voxel                 & -                     & 65.3                      & 56.9                      & 0.285                         & 0.253                     & 0.323                     & 0.272                     & 0.186 \\
        \midrule
        RCM-Fusion \cite{kim2024rcm}                        & C+R                 & R50                   & $450\times800$        & 52.9                      & 44.3                      & -                       & -                     & -                    & -                     & - \\
        X3KD \cite{klingner2023x3kd}                        & C+R               & R50                   & $256\times704$        & 53.8                      & 42.3                      & -                             & -                         & -                         & -                         & - \\
        StreamPETR \cite{Wang2023streampetr}                & C                 & R50                   & $256\times704$        & 54.0                      & 43.2                      & 0.581                         & 0.272                     & 0.413                     & 0.295                     & 0.195 \\
        BEVNeXt \cite{li2024bevnext}                        & C                 & R50                   & $256\times704$        & 56.0                      & 45.6                      & 0.530                         & 0.264                     & 0.424                     & 0.252                     & 0.206 \\
        CRN \cite{kim2023crn}                               & C+R               & R50                   & $256\times704$        & 56.0                      & 49.0                      & 0.487                         & 0.277                     & 0.542                     & 0.344                     & 0.197 \\
        RCBEVDet \cite{lin2024rcbevdet}                     & C+R               & R50                   & $256\times704$        & 56.8                      & 45.3                      & 0.486                         & 0.285                     & 0.404                     & 0.220                     & 0.192 \\
        HyDRa \cite{wolters2024hydra}                       & C+R               & R50                   & $256\times704$        & 58.5                      & 49.4                      & 0.463                         & 0.268                    & 0.478                     & 0.227            & 0.182 \\
        \gr \textbf{\sotacolor{\name}}                             & C+R               & R50                   & $256\times704$        & \textbf{62.0}             & \textbf{54.5}             & 0.496                         & 0.269                    & 0.403                     & 0.177            & 0.181 \\
    
        \midrule
        MVFusion \cite{wu2023mvfusion}                      & C+R               & R101                  & $900\times1600$       & 45.5                      & 38.0                      & 0.675                         & 0.258                     & 0.372                     & 0.833                     & 0.196 \\
        FUTR3D \cite{chen2022futr3d}                        & C+R               & R101                  & $900\times1600$       & 50.8                      & 39.9                      & -                             & -                         & -                         & 0.561                     & - \\ %
        SparseBEV \cite{liu2023sparsebev}       & C                 & R101                  & $512\times1408$       & 59.2                      & 50.1                      & 0.562                         & 0.265                     & 0.321                     & 0.243                     & 0.195 \\ %
        StreamPETR \cite{Wang2023streampetr}    & C                 & R101                  & $512\times1408$       & 59.2                      & 50.4                      & 0.569                         & 0.262                     & 0.315            & 0.257                     & 0.199 \\
        CRN \cite{kim2023crn}                               & C+R               & R101                  & $512\times1408$       & 59.2                      & 52.5                      & 0.460                         & 0.273                     & 0.443                     & 0.352                     & 0.180 \\
        Far3D \cite{jiang2024far3d}                         & C                 & R101                  & $512\times1408$       & 59.4                      & 51.0                      & 0.551                         & 0.258                     & 0.372                     & 0.238                      & 0.195 \\
        BEVNeXt \cite{li2024bevnext}                        & C                 & R101                  & $512\times1408$       & 59.7                      & 50.0                      & 0.487                         & 0.260                     & 0.343                     & 0.245                      & 0.197\\
        HyDRa \cite{wolters2024hydra}                     & C+R               & R101                  & $512\times1408$       & 61.7             & 53.6             & 0.416                & 0.264            & 0.407                     & 0.231            & 0.186 \\
        \gr \textbf{\sotacolor{\name}}                             & C+R               & R101                   & $512\times1408$        & \textbf{64.4}             & \textbf{57.1}             & 0.484                        & 0.264                    & 0.308                     & 0.175            & 0.178 \\
        
        \bottomrule
    \end{tabular}
    }
    \caption{
        \textbf{3D Object Detection} on nuScenes \texttt{val} set. 
        `L', `C', and `R' represent LiDAR, Camera, and Radar, respectively.
    }
\label{table:nusc_val}
\end{table*}

%% file: sec/4_experiments.tex
\section{Experiments}
\label{sec:experiments}
\subsection{Datasets}

We use the two large-scale radar datasets, nuScenes \cite{caesar2020nuscenes} and the new TruckScenes \cite{fent2024man} to explore, generalize, and validate the findings on our \name~model architecture.

The CVPR \textbf{nuScenes} dataset \cite{caesar2020nuscenes} is the traditional research benchmark for radar-fusion-based 3D perception. 
In the urban scenario of Boston and Singapore, 1000 scenes of 20s are captured with six cameras, five radar, and one LiDAR sensor. The annotation range is 50 m.

Recently, the NeurIPS \textbf{TruckScenes} \cite{fent2024man} benchmark was introduced to provide high-quality and modern 4D radar point clouds and diverse scenes for autonomous trucking.
Four cameras, six LiDAR, and six 4D imaging radar sensors capture 740 scenes of 20s in 360 degree coverage. 
The biggest differentiation is the annotation range of 150 m, dynamic faster speeds of highway driving, and diversity of adverse splits, making it challenging for single-modal perception systems.

\input{tables/nusc_test}
\input{tables/nusc_tracking}
\input{tables/trusc_val}

\subsection{Evaluation Metrics}
Both benchmarks follow the official evaluation metrics of the nuScenes Detection Score (NDS).
It comprises the weighted sum of the mean Average Precision (mAP) and five True Positive metrics: Translation (mATE), Scale (mASE), Orientation (mAOE), Velocity (mAVE), and Attribute Error (mAAE).
Further details can be found in \cite{caesar2020nuscenes}.
For evaluating multi-object tracking performance, we use the official normalized AMOTA metric \cite{weng2019baseline}, trading off false positives, missed targets, and identity switches.

\subsection{Implementation Details} 
We adopt StreamPETR \cite{Wang2023streampetr} and Far3D\cite{jiang2024far3d} as our baseline for the camera stream and follow standard practices for training and hyperparameters \cite{Wang2023streampetr}.
For a fair comparison, we employ pretrained ResNet \cite{he2016deep} and V2-99 \cite{lee2019vov} backbone encoders.
For the radar stream, we utilize a downscaled PointTransformerV3 \cite{wu2024point} with randomly initialized weights on multiple sweeps of radar, on RCS and Doppler features. 
We use the 16 closest radar points in the frustum space 
whereas the LSA module leverages a local neighborhood of 32 queries (instead of the default setting of 644 normal + 256 temporal~+~600 denoising queries \cite{jiang2024far3d, Wang2023streampetr}).

Due to sequential sampling \cite{park2022time,Wang2023streampetr} we train for 24 epochs in ablations and 60 when comparing with others, trading off step-to-step sampling diversity for memory consumption.
Like CRN \cite{kim2023crn} and RCBEVDet \cite{lin2024rcbevdet}, the inference time is measured with a single batch and FP16 precision on an RTX3090 GPU.
No test-time augmentations, CBGS~\cite{zhu2019class} or future frames are used.

\subsection{Main Results}
We compare \name~to the previous state-of-the-art methods on the val and test sets of nuScenes and TruckScenes.

\noindent \textbf{nuScenes Val.}
As reported in \cref{table:nusc_val}, \name~consistently outperforms both BEV-based as well as query-based methods in terms of NDS and mAP (+5.1 for R50 and +3.5 for R101).
Notably, the object-level motion modeling can benefit greatly from the Doppler velocity, as shown by the large improvement in mAVE.
Especially in the small scale and low resolution (real-time) scenarios, the performance gain over the vision-based methods is significant.

\noindent \textbf{nuScenes Test.}
When scaling up to the V2-99 backbone and evaluating on the test server (\cf \cref{table:nusc_test} and \cref{table:nusc_tracking}), \name~introduces a new state-of-the-art in 3D object detection on nuScenes, with an NDS of 67.1 (+2.9) and mAP of 60.0 (+2.6), surpassing the previous best camera- or radar-based methods.
Capitalizing on the high accuracy, strong motion modeling and paired with a velocity-based greedy tracker \cite{yin2021center}, \name~achieves also the best tracking-by-detection performance, increasing the AMOTA to 63.1~(+4.7).

\noindent \textbf{TruckScenes Val.}
Our architecture generalizes well to the new domain and sensor setup of TruckScenes, achieving a competitive NDS of 37.4 on the validation set (\cf \cref{table:trusc_val}).
With more adverse conditions and longer detection ranges of up to 150 meters, the adoption of radar becomes more important, as \name~doubles the mAP (+11.8) over the current vision-only state-of-the-art.
We provide more detailed information about the ranges, conditions, and classes in the Appendix.

\input{tables/abl_combined}
\noindent \textbf{TruckScenes Test.}
\name~surpasses all single-modal baselines and achieves competitive mAP scores to the LiDAR model (\cf. \cref{table:trusc_val}).
We set the state-of-the-art for vision-centric and radar-based methods on TruckScenes, with a long-range NDS score of 37.4, demonstrating the effectiveness of our object-centric architecture and the inclusion of 4D radar.
Moreover, we emphasize the importance of Doppler information in high-speed and dynamic scenarios. 
While radar-only methods struggle with overall detection performance, camera-radar fusion effectively leverages Doppler measurements to achieve strong velocity prediction accuracy.

\begin{figure*}[!htpb]
    \centering
    \includegraphics[width=1.0\textwidth]{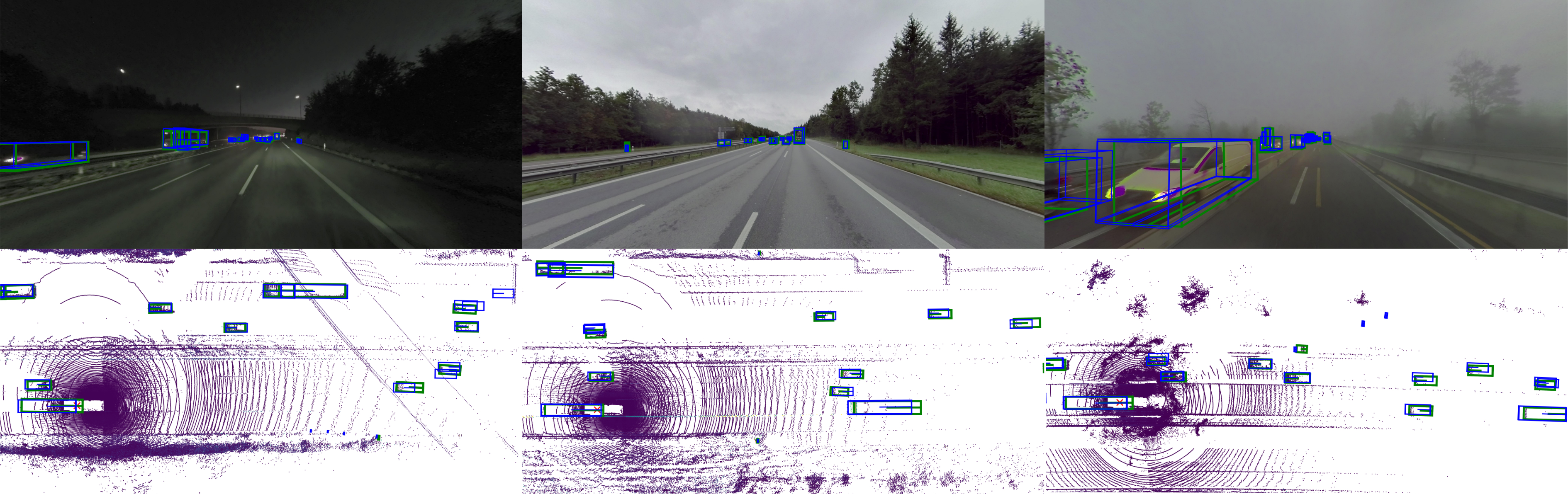}
    \caption{
        \textbf{Qualitative Results} of SpaRC performing in challenging conditions (night, long distance, fog) of the TruckScenes validation set. We visualize the front camera and the corresponding top-down view of the driving corridor with up to 150 m of detection range. Predictions are visualized with blue bounding boxes and annotations in green. LiDAR is displayed for geometric context. %
        }
    \label{fig:qualitative}
\end{figure*}
\subsection{Ablation Studies}
\noindent \textbf{Component Analysis.} 
We conduct extensive ablation studies to analyze the effectiveness of each component in our model. 
As shown in \cref{table:nusc_ablations}, we start from a camera-only baseline using Far3D \cite{jiang2024far3d} trained for 24 epochs. 
Adding our Local Self-Attention (LSA) module improves performance by 1.2 NDS and 1.4 mAP, demonstrating the benefits of focusing attention on locally relevant features even without radar input. 
The efficiency gains from LSA enable incorporating additional fusion components.
Introducing the Range-Adaptive Radar (RAR) module yields substantial improvements of 4.1 NDS and 5.9 mAP. 
By leveraging Doppler velocity information through motion-aware layer normalization, RAR reduces velocity errors (mAVE) by 34\% while also improving localization accuracy (mATE). 
Finally, the Sparse Frustum Fusion (SFF) module further boosts performance by 1.0 NDS and 1.6 mAP by effectively incorporating radar point features in the perspective view.
The ablation results validate that each component contributes meaningfully to the final performance, with the full model achieving significant gains of 6.3 NDS and 8.9 mAP over the camera-only baseline. 
This demonstrates the effectiveness of our sparse fusion design in leveraging complementary radar information.

In \cref{table:nusc_rar}, we analyze the impact of the range guidance of the RAR module. When excluding the range-modulation (-2.9 mAP) or the hierarchical point structuring~(-1.9 mAP), both components show a significant decrease in performance.
The best performance is achieved when including the locality bias and point-based modeling.

\noindent \textbf{Achieving Real-time Speed.} 
We analyze the inference latency vs. performance trade-off in \cref{fig:inference_analysis} across backbone architectures (R18, R50, R101). The results demonstrate that \name~achieves superior accuracy at lower computational cost compared to existing methods, validating the efficiency advantages of our sparse object-centric fusion.

\subsection{Qualitative Results}
\cref{fig:qualitative} demonstrates the robustness of our method by presenting detection results in three challenging highway scenarios, including nighttime driving, long-range detection, and adverse weather conditions such as fog.
\input{tables/abl_rar}

\subsection{Limitations and Outlook}
\label{sec:outlook}

For now, we rely on pre-processed radar data in the form of point clouds. 
However, lower-level radar representations like the high-dimensional radar cube also pose a promising research direction, with new challenges to efficient computations \cite{paek2022k, fent2024dpft,ding2024radarocc}. %
In future work, we want to explore how SpaRC and its sparse modeling can handle dense and raw radar tensors.

%% file: tables/nusc_test.tex
\setlength{\tabcolsep}{2mm}
\begin{table}[!t]
\centering
\resizebox{\linewidth}{!}{
\begin{tabular}{l|c|c|cc} 
    \toprule
    \textbf{Methods} &                          \textbf{Input}      & \textbf{Backbone}         & \textbf{NDS$\uparrow$}    & \textbf{mAP$\uparrow$}  \\
    \midrule
    PointPillars \cite{lang2019pointpillars}    & L                 & Pillars                   & 55.0                      & 40.1                     \\ %
    CenterPoint \cite{yin2021center}            & L                 & Voxel                     & 67.3                      & 60.3                      \\
    \midrule
    KPConvPillars \cite{ulrich2022impr}         & R                 & Pillars                   & 13.9                      &  4.9                      \\
    RadarDistill \cite{bang2024radardistill}    & R                 & Pillars                   & 43.7                      & 20.5 \\
    \midrule
    CenterFusion \cite{nabati2021centerfusion}  &C+R                & DLA34                     & 44.9                      & 32.6                    \\
    MVFusion \cite{wu2023mvfusion}              &C+R                & V2-99                     & 51.7                      & 45.3                  \\
    CRAFT \cite{kim2022craft}                   &C+R                & DLA34                     & 52.3                      & 41.1                    \\
    BEVDepth \cite{li2022bevdepth}              & C                 & ConvNeXt-B                & 60.9                      & 52.0                     \\
    SOLOFusion \cite{park2022time}              & C                 & ConvNeXt-B                & 61.9                      & 54.0                      \\
    BEVFormerV2 \cite{yang2023bevformerV2}      & C                 & InternImage-B             & 62.0                      & 54.0                  \\
    CRN \cite{kim2023crn}                       &C+R                & ConvNeXt-B                & 62.4                      & 57.5                 \\
    StreamPETR \cite{Wang2023streampetr}        & C                 & V2-99                     & 63.6                      & 55.0                     \\
    SparseBEV \cite{liu2023sparsebev}           & C                 & V2-99                     & 63.6                      & 55.6                  \\    
    RCBEVDet \cite{lin2024rcbevdet}             & C+R               & V2-99                     & 63.9                      & 55.0 \\
    HyDRa \cite{wolters2024hydra}                &C+R                & V2-99                     & 64.2             & 57.4            \\
    \gr \textbf{\sotacolor{\name}}                        &C+R                & V2-99                     & \textbf{67.1}             & \textbf{60.0}           \\ %
    \bottomrule
\end{tabular}
}
\caption{
    \textbf{3D Object Detection} on the nuScenes \texttt{test} set.
}\label{table:nusc_test}
\end{table}

%% file: tables/nusc_tracking.tex
\renewcommand{\thefootnote}{2}
\setlength{\tabcolsep}{0.3em}
\begin{table}[!t]
\centering
\resizebox{\columnwidth}{!}{
\begin{tabular}{l|c|c|cc} %
    \toprule
    \textbf{Methods}                                        & \textbf{Input}    & \textbf{Backbone}     & \textbf{AMOTA$\uparrow$}  & \textbf{AMOTP$\downarrow$}   \\ %
    \midrule
    CenterPoint \cite{yin2021center}                        & L                 & Voxel                 & 63.8                      & 0.555                         \\ %
    \midrule
    UVTR \cite{li2022unifying}                              & C                 & \small{V2-99}         & 51.9                      & 1.125                         \\ %
    ByteTrackV2 \cite{zhang2023bytetrackv2}                 & C                 & \small{V2-99}         & 56.4                      & 1.005                         \\ %
    StreamPETR \cite{yang2022quality}                       & C                 & \small{ConvNeXt-B}    & 56.6                      & 0.975                         \\ %
    CRN  \cite{kim2023crn}                                  & C+R               & \small{ConvNeXt-B}    & 56.9                      & \textbf{0.809}                         \\ %
    HyDRa                                                   & C+R               & \small{V2-99}         & 58.4                      & 0.950                         \\ %
    \gr \textbf{\sotacolor{\name}}                                      & C+R               & \small{V2-99}         & \textbf{63.1}             & 0.901                         \\ %
    \bottomrule
\end{tabular}
}
\caption{
    \textbf{3D Object Tracking} on nuScenes \texttt{test} set.
}\label{table:nusc_tracking}
\end{table}

%% file: tables/trusc_val.tex
\begin{table*}[!t]
    \centering
    \resizebox{0.9\textwidth}{!}{
    \begin{tabular}{l|c|c|c|c|cc|c@{\hspace{1.0\tabcolsep}}c@{\hspace{1.0\tabcolsep}}c@{\hspace{1.0\tabcolsep}}c@{\hspace{1.0\tabcolsep}}c} 
        \toprule
        \textbf{Methods}                                    & \textbf{Input}    & \textbf{Backbone}     & \textbf{Image Size}  & \textbf{Split}  & \textbf{NDS$\uparrow$}    & \textbf{mAP$\uparrow$}    & \textbf{mATE$\downarrow$}     & \textbf{mASE$\downarrow$} & \textbf{mAOE$\downarrow$} & \textbf{mAVE$\downarrow$} & \textbf{mAAE$\downarrow$} \\
        
        \midrule
        CenterPoint-V$^*$ \cite{yin2021center}                  & L                 & Voxel                 & -                    & \texttt{val}                 & 35.3                      & 22.6                      & 0.461                         & 0.405                     & 0.468                     & 3.028                     & 0.261 \\
        \midrule
        Far3D \cite{jiang2024far3d} & C & V2-99                  & $640\times960$        & \texttt{val} & 21.4 & 10.7 & 0.883 & 0.507 & 0.671 & 1.352 & 0.338  \\
        
        \gr\textbf{\sotacolor{\name} }                                  & C+R               & V2-99                  & $640\times960$                        & \texttt{val} & \textbf{35.4} & \textbf{22.5} & 0.798                        & 0.449                    & 0.476                     & 0.613            & 0.248 \\
        
        \midrule
        CenterPoint-V$^*$ \cite{yin2021center}                  & L                 & Voxel                 & -                          & \texttt{test}                    & 41.0 & 26.7                 & 0.409                     & 0.352                   & 0.277                 & 2.730                     & 0.201 \\
        \midrule
        RadarGNN$^*$ \cite{fent2023radargnn}                    & R                 & -                     & -                 & \texttt{test}                                                                  & 10.7 & 7.0 & 0.892 & 0.809 & 1.132 & 8.003 & 0.571 \\
        PETR$^*$ \cite{liu2022petr}                             & C                 & V2-99                 & $300\times800$ & \texttt{test}          & 12.1 & 2.2 &1.125 & 0.686 & 0.647 & 1.499 & 0.564 \\
        
        HyDRa \cite{wolters2024hydra}                     & C+R               & V2-99                  & $928\times1952$ & \texttt{test} & 22.4 & 12.8 & 0.725 & 0.544 & 0.744 & 1.180 &	0.388 \\
        \textbf{\sotacolor{\name}}                                 & C+R               & V2-99                  & $928\times1952$      & \texttt{test}                         & \textbf{37.4} & \textbf{27.2} & 0.759 & 0.413 & 0.411 & 0.814 & 0.227 \\
        \midrule
        \gr\textbf{\sotacolor{\name}} & C+R               & Intern-Image L                  & $928\times1952$ & \texttt{test} & \textbf{42.0} & \textbf{33.2} & 0.717 & 0.393 & 0.355 & 0.784 & 0.201 \\

        \bottomrule
    \end{tabular}
    }
    \caption{
        \textbf{3D Object Detection} on TruckScenes \texttt{val} and \texttt{test} sets. The detection range is up to 150 m ($^*$denotes the official baselines).
    }
\label{table:trusc_val}
\end{table*}

%% file: tables/abl_combined.tex
\begin{table}[!t]
    \centering
    \resizebox{\linewidth}{!}{
    \setlength{\tabcolsep}{2mm}
    \begin{tabular}{lcc|c|cc|cc}
        \toprule
        \textbf{SFF} &  \textbf{RAR} & \textbf{LSA} & \textbf{Input}        &\textbf{NDS$\uparrow$}    &\textbf{mAP$\uparrow$}       & \textbf{mATE$\downarrow$} & \textbf{mAVE$\downarrow$} \\ %
        \midrule
        & &                       & C   & 48.6 & 37.5 & 0.672 & 0.309 \\
        \midrule
        \cmark &  &                     & C+R & 51.4                    & 40.7                      & 0.644 & 0.252 \\
         & \cmark &                     & C+R & 53.5                    & 43.8                       & 0.587 & 0.199 \\
         &  & \cmark                    & C   & 49.8                    & 38.9                       & 0.645 & 0.301 \\ 
         & \cmark & \cmark                    & C+R & 53.9                    & 44.8                       & 0.590 & 0.198 \\ 
        \cmark & \cmark &                     & C+R & 54.5                    & 45.1                       & 0.581 & 0.205 \\
        \gr \cmark & \cmark & \cmark                & C+R & \textbf{54.9}           & \textbf{46.4}              & \textbf{0.580} & \textbf{0.195} \\
        \bottomrule
    \end{tabular}
    }
    \vspace{-0.2cm}
    \caption{
    \textbf{Ablation} of SpaRC components on nuScenes \texttt{val} set
    LSA - Local Self-Attention;
    RAR - Range Adaptive Radar Aggregation;
    SFF - Sparse Frustum Fusion.
    }
    \label{table:nusc_ablations}
\end{table}

%% file: tables/abl_rar.tex
\begin{table}[!ht]
    \centering
    \resizebox{\linewidth}{!}{
    \setlength{\tabcolsep}{2mm}
    \begin{tabular}{l|c|c c|c c}
        \toprule
        \textbf{Methods}                            & \textbf{Input}        &\textbf{NDS$\uparrow$}    &\textbf{mAP$\uparrow$}       & \textbf{mATE$\downarrow$}   & \textbf{mAVE$\downarrow$} \\
        \midrule
        Camera Baseline \cite{jiang2024far3d}                              & C                     & 48.6                    & 37.5                       & 0.672                       & 0.309    \\
        SpaRC RAR V1                                      & C+R                     & 52.3                    & 41.9                       & 0.612                       & 0.209    \\
        SpaRC RAR V2                                 & C+R     & 52.3 & 42.9 & 0.608 & 0.246                   \\
        \gr SpaRC RAR V3                         & C+R                   & \textbf{53.9}           & \textbf{44.8}              & \textbf{0.590}                 & \textbf{0.198}    \\
        \bottomrule
    \end{tabular}
    }
    \caption{
    \textbf{Ablation} variants of SpaRC Range Adaptive Radar Aggregation (RAR) module on nuScenes \texttt{val} set
    (V1 - Radar aggregation without focused range guidance; V2 - RAR without serialized point encoding; V3 - the full RAR module).
    }
    \label{table:nusc_rar}
    \end{table}

%% file: sec/6_conclusion.tex
\section{Conclusion}
\label{sec:conclusion}

In this paper, we introduce \textbf{SpaRC}, a novel camera-radar fusion transformer for 3D object detection. 
We overcome the limitations of dense BEV-based fusion methods and address key challenges of query-based architectures.
By introducing sparse but strong local cues into the decoder, we concentrate computational resources on salient regions of the radar modality and reduce the uncertainty of implicit 3D decoding.
This information-rich but sparse representation achieves superior performance in accuracy and robustness over all existing vision-centric and radar-based methods.
We achieve a new state-of-the-art in camera-radar fusion on the nuScenes and TruckScenes benchmarks in both short-range urban environments and dynamic long-range high-way scenarios.
Our real-time capable architecture provides an efficient and scalable solution for autonomous driving perception, bridging the gap to LiDAR-centric methods.

%% file: sec/7_suppl.tex
\clearpage
\setcounter{section}{0}
\maketitlesupplementary

\def\thesection{\Alph{section}}

\section{Overview}
\label{sec:overview}

This supplementary material presents additional technical details and experimental analysis of our proposed method, SpaRC. 
We provide in-depth architectural specifications, ablation studies, and both quantitative and qualitative results to validate our approach.
We begin with implementation specifications of the SpaRC architecture, including backbone design choices, training protocols, and inference optimizations (\cref{sec:implementation}). 
Subsequently, we conduct ablation studies focusing on challenging scenarios in the nuScenes and TruckScenes datasets, particularly examining performance under adverse weather and lighting conditions (\cref{sec:ablation}).
To conclude, we present a qualitative analysis comparing our multi-modal fusion approach with camera-only baselines, demonstrating the effectiveness of our method across diverse real-world scenarios (\cref{sec:qualitative}).

\section{Implementation Details}
\label{sec:implementation}

\noindent \textbf{Radar Point Backbone.}
Our radar point processing backbone is based on the Point Transformer architecture \cite{zhao2021point, wu2024point}, which encodes radar point clouds through a structured serialization format. 
This approach preserves individual point representations while achieving permutation-invariance without requiring sparse BEV-grid representations. 
We employ space-filling curves and local neighborhood mapping to facilitate efficient point patch grouping, enabling the subsequent application of self-attention for feature extraction while maintaining locality-preserving properties.

The architecture follows a hierarchical U-Net-style encoder-decoder structure with four stages, enabling multi-scale feature extraction.
Preserving the hierarchical structure, we reduce the model capacity by decreasing the depth of encoder and decoder layers and adjusting embedding dimensions and patch sizes.
This adaptation accounts for the relatively lower point cloud density of radar data compared to typical LiDAR applications.
We operate at a sample resolution of 0.05 m with a patch size of 64.
This design enables efficient processing of radar point clouds while maintaining real-time performance requirements for autonomous perception systems.
The architectural specifications, including encoder-decoder configurations, are summarized in \cref{table:ptv3_settings}.

\input{tables/suppl/ptv3_settings}
\input{tables/suppl/nusc_inference}

\noindent \textbf{Training Details.}
Our model architecture employs 6 decoder layers with a local query neighborhood of 32 and samples the 16 closest radar points in frustum space. For ablation studies, we train the model for 24 epochs, while full experimental comparisons utilize 60 epochs of training.

The standard model configuration utilizes 644 object queries \cite{Wang2023streampetr, jiang2024far3d}. 
For TruckScenes, we linearly scale the number of queries to 1800 to account for the thrice extended detection range rather than quadratically increasing the BEV space representation. 
Following established practice \cite{lin2024rcbevdet,wolters2024hydra,Wang2023streampetr,jiang2024far3d}, we employ a VoVNet-99 backbone \cite{lee2019vov} for the camera stream, pretrained with FCOS3D \cite{wang2021fcos3d} on the nuScenes dataset.
During training, we apply both perspective and 3D data augmentations. 
These include image resizing, random cropping, rotation, and horizontal flipping, with consistent augmentations applied to radar data in the perspective view. 
The perception range spans [-51.2, 51.2] meters for nuScenes and [-152.4, 152.4] meters for TruckScenes.
We optimize the model using AdamW \cite{loshchilov2017decoupled} with a learning rate of 4e-4, cosine annealing policy, weight decay coefficient of 0.05, and a batch size of 16. 

\input{tables/suppl/trusc_classes}

We maintain consistency with Far3D and ArgoVerse \cite{wilson2023argoverse} conventions for long-range detection scenarios by utilizing the same nuScenes-pretrained VoV-99 backbone for TruckScenes.
We emphasize that our method does not rely on test-time augmentations, class-balanced grouping and sampling (CBGS)~\cite{zhu2019class}, or future frame information.
Comprehensive training configurations will be made publicly available in our code repository.

\noindent \textbf{Inference Optimization.}
We propose four model configurations (SpaRC Tiny, Small, Base, and Large) that achieve state-of-the-art performance compared to existing real-time approaches \cite{kim2023crn,lin2024rcbevdet,Wang2023streampetr} on the nuScenes benchmark.
Through extensive experimentation, we demonstrate that our expressive fusion features enables a reduction of model size and complexity.
Specifically, we optimize the model by removing the perspective head during inference, eliminating dynamic query allocation from the perspective 3D head \cite{jiang2024far3d}, and parallelizing backbone networks using CUDA streams for efficient GPU utilization.
The most impactful parameters affecting inference speed are the number of decoder layers, query count, and depth of radar encoding encoder-decoder blocks. 
The Tiny configuration drops the point serialization for a single hierarchical downsampling aggregation of direct neighboring points \cite{xu2023frnet}.
We summarize the main parameters in \cref{table:nusc_inference}, utilizing a two-level feature pyramid with strides of 16 and 32.
We will provide the pre-trained models in the accompanying code repository.

In the breakdown of Fig.~\ref{fig:sparc_latency_breakdown}, we show that our sparse approach introduces minimal computational overhead while achieving significant performance gains. 

\section{Additional Ablations}
\label{sec:ablation}

\noindent \textbf{LSA Configuration.} Our ablation study (\cref{table:lsa}) on LSA demonstrates that K=32 (of all 644 queries) provides optimal performance, suggesting that focused local spatial context is sufficient for 3D object detection.

\noindent \textbf{Range Analysis.}
\cref{table:trusc_range} presents a range analysis of SpaRC compared to the baseline Far3D on the TruckScenes validation set.
The performance gap between SpaRC and Far3D widens as the detection range increases from 25 to 150 meters, with SpaRC maintaining strong detection capabilities even at longer distances where Far3D's performance degrades more significantly.
At the 50-meter range, SpaRC achieves a 44\% improvement in mAP over the baseline. 
When evaluating across the full 150-meter detection range, SpaRC more than doubles the mAP score compared to Far3D, demonstrating superior long-range detection capabilities.
This underscores the critical importance of radar fusion for robust localization in 3D perception.  

\input{tables/suppl/trusc_range}

\input{tables/suppl/nusc_weather}

\input{tables/abl_lsa}
\begin{center}
    \centering
    \includegraphics[width=\columnwidth]{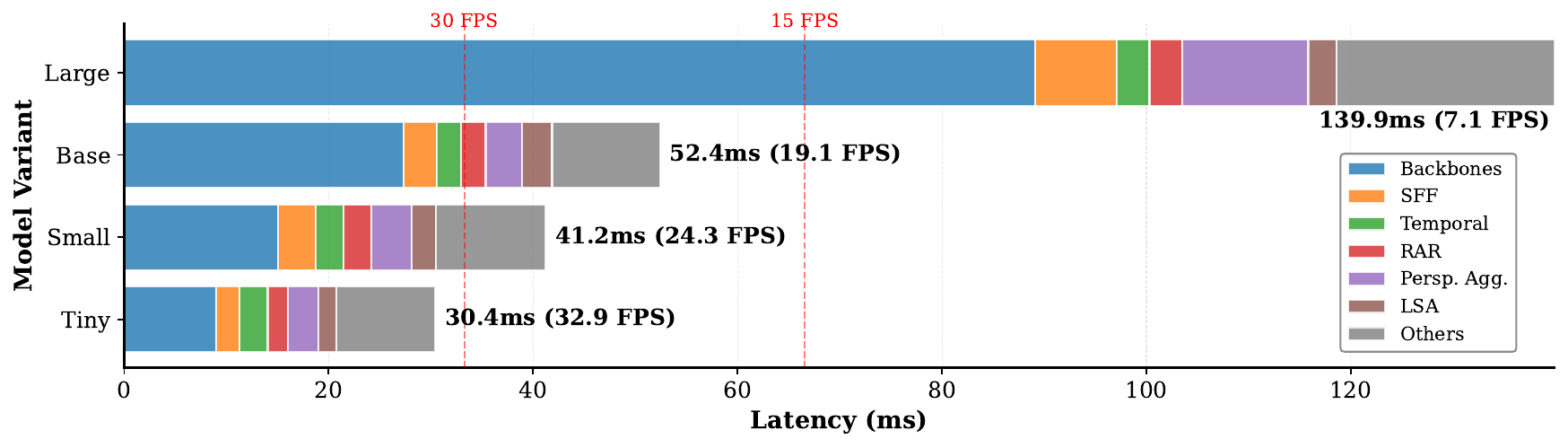}
    \captionof{figure}{
        \textbf{Latency breakdown} of SpaRC model variants.
    }
    \label{fig:sparc_latency_breakdown}
\end{center}

\noindent \textbf{Per Class Analysis.}
For a holistic understanding of SpaRC's performance across different object classes (\cf \cref{table:trusc_classes}), we present per-class mAP scores in \cref{table:trusc_classes}.
The radar-derived metric measurements significantly enhance the detection capabilities of our approach, enabling robust identification of both small-scale objects such as vulnerable road users (pedestrians +8.9 mAP, cyclists +5.2 MAP, motorcycles +11.8 mAP), which traditionally pose challenges for camera-only methods, as well as highly reflective metallic objects, including cars (+27.2 mAP) and trucks (+17.5 mAP). 
This demonstrates the complementary nature of the multi-modal fusion approach for general robust depth estimation.

\input{tables/suppl/trusc_weather}
\noindent \textbf{Adverse Conditions.}
Next, we analyze SpaRC's performance across diverse adverse environmental conditions. 
On the nuScenes dataset (\cf \cref{table:nusc_weather}), our radar-based approach demonstrates particularly large improvements in low-light scenarios, achieving a +5.7 mAP gain.
Notably, our method shows superior robustness in challenging lighting conditions compared to previous BEV-grid approaches like CRN \cite{kim2023crn}, validating the effectiveness of our adaptive fusion mechanism in handling adverse scenarios.

The performance advantages become even more pronounced when evaluating on the TruckScenes dataset (\cf \cref{table:trusc_weather}), which presents more challenging driving situations and higher-quality 4D radar measurements. 
Our analysis reveals that SpaRC's radar-enhanced perception provides consistent improvements across varying environmental conditions, including different times of day, seasons, and lighting conditions.
The high-fidelity radar information proves especially beneficial for long-range detection scenarios, where camera-only approaches struggle with accurate depth estimation and localization even under favorable conditions like clear daylight.

This demonstrates the robustness of our adaptive fusion approach, which effectively leverages complementary radar information when visual features become less reliable.
The consistent performance across diverse environmental conditions validates the generalizability of our multi-modal architecture and its capability to maintain reliable perception even in challenging real-world scenarios.

\noindent \textbf{Robustness Analysis}
Unlike methods relying on exact geometric projections (e.g. occupancy-based CRN), SpaRC handles misalignment through learnable positional embeddings and soft attention mechanisms. Our SFF uses
column-wise topK attention rather than direct point-to-pixel projection, while RAR dynamically adjusts feature interactions based on learned spatial relationships. This learning-based approach inherently adapts to both spatial and temporal misalignments. \cref{fig:sensor_misalignment} demonstrates SpaRC’s superior robustness to calibration errors (adding Gaussian noise to radar points), significantly outperforming HyDRa \cite{wolters2024hydra} and CRN \cite{kim2023crn} across all noise levels.

\section{Qualitative Comparison}
\label{sec:qualitative}
In this section, we conduct a comprehensive qualitative evaluation comparing SpaRC with Far3D, demonstrating the efficacy of our approach in mitigating depth uncertainty and reducing false positive detections along projection rays in 3D space.

\begin{center}
    \centering
    \includegraphics[width=\columnwidth]{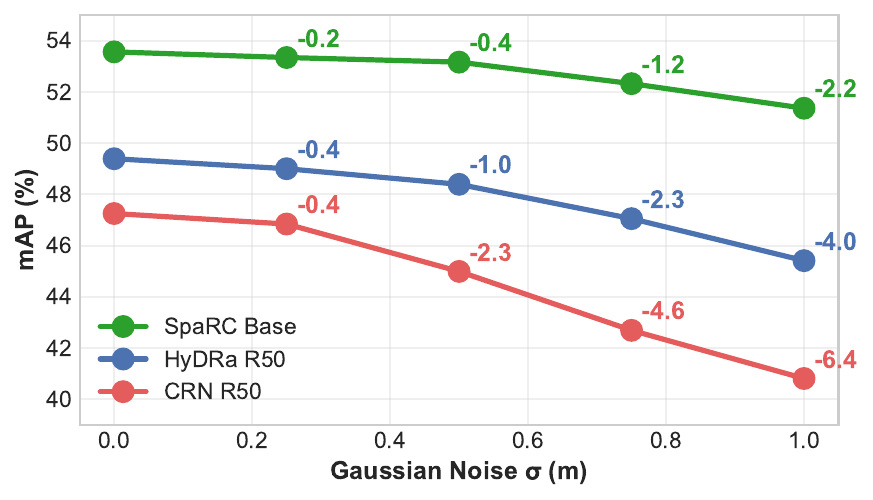}
    \captionof{figure}{
        Robustness of mAP under different levels of radar misalignment (gaussian noise with std. $\sigma$) on nuScenes \texttt{val} set.
        }
    \label{fig:sensor_misalignment}
\end{center}

We present qualitative results across diverse and challenging scenarios from the TruckScenes validation set. 
We incorporate both front-view camera images and corresponding bird's-eye-view representations of the driving corridor, extending to a maximum detection range of 150 meters. 
Object detections are denoted by blue bounding boxes, while ground truth annotations are rendered in green. 
For geometric context and spatial reference, we overlay LiDAR point cloud data.

\begin{figure*}[htbp]
    \centering
    \includegraphics[width=0.82\textwidth]{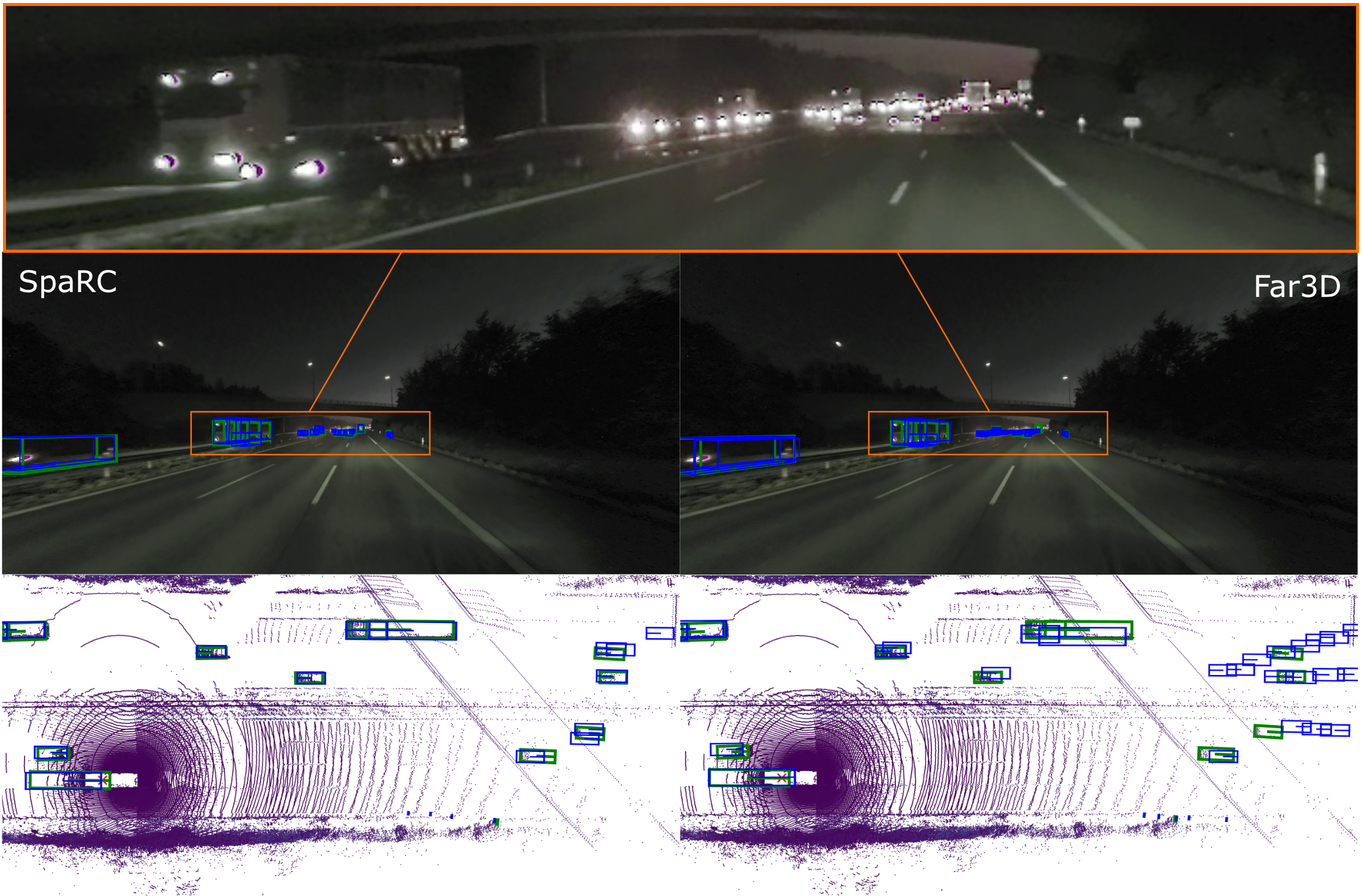}
    \caption{
        \textbf{Qualitative Comparison of SpaRC and Far3D: Night Scenario.} 
        We visualize predictions (blue) vs ground truth (green) of 3D object detection on the TruckScenes val set in the perspective and top-down view (150 m detection range). 
        SpaRC is visualized on the left and Far3D on the right. 
        The top row shows a zoomed-in cutout of the long-range detection region for better visibility. 
    }
    \label{fig:comp_1}
\end{figure*}

\begin{figure*}[htbp]
    \centering
    \includegraphics[width=0.82\textwidth]{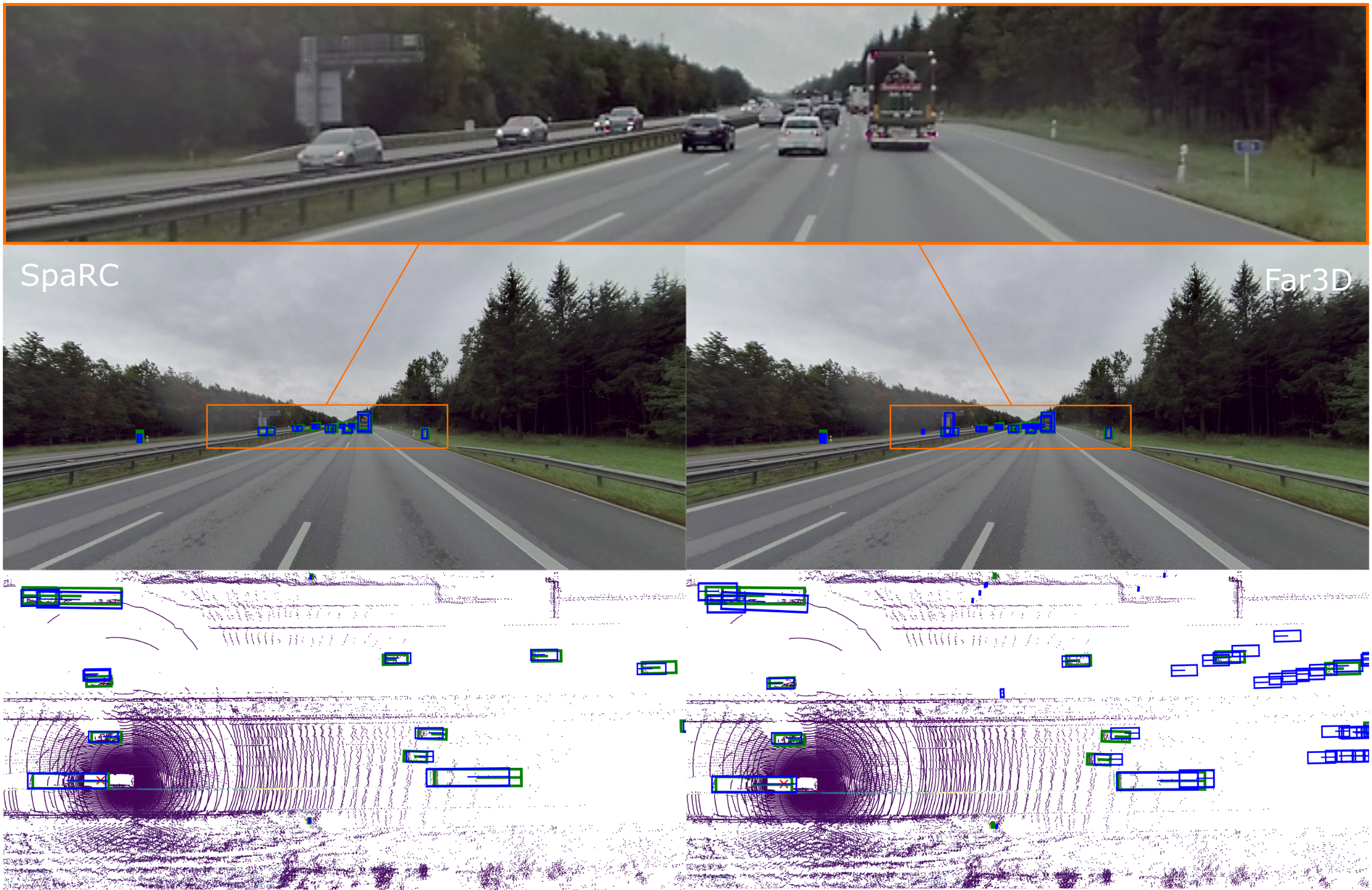}
    \caption{
        \textbf{Qualitative Comparison of SpaRC and Far3D: Overcast and Long Distance.} 
        We visualize predictions (blue) vs ground truth (green) of 3D object detection on the TruckScenes val set in the perspective and top-down view (150 m detection range). 
        SpaRC is visualized on the left and Far3D on the right. 
        The top row shows a zoomed-in cutout of the long-range detection region for better visibility. 
    }
    \label{fig:comp_5}
\end{figure*}

\begin{figure*}[htbp]
    \centering
    \includegraphics[width=0.82\textwidth]{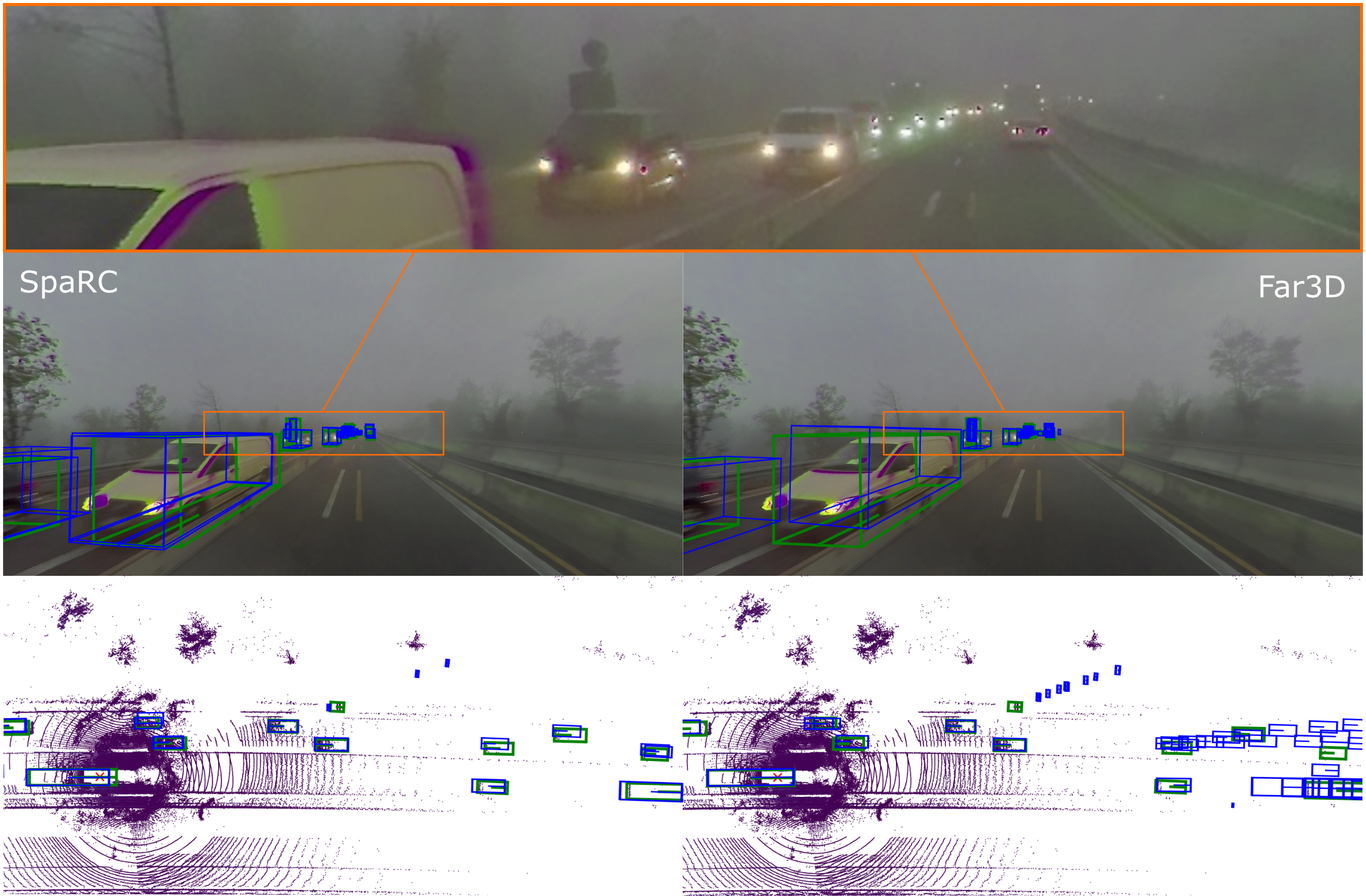}
    \caption{
        \textbf{Qualitative Comparison of SpaRC and Far3D: Foggy Scene } 
        We visualize predictions (blue) vs ground truth (green) of 3D object detection on the TruckScenes val set in the perspective and top-down view (150 m detection range). 
        SpaRC is visualized on the left and Far3D on the right. 
        The top row shows a zoomed-in cutout of the long-range detection region for better visibility. 
    }
    \label{fig:comp_4}
\end{figure*}

\begin{figure*}[htbp]
    \centering
    \includegraphics[width=0.82\textwidth]{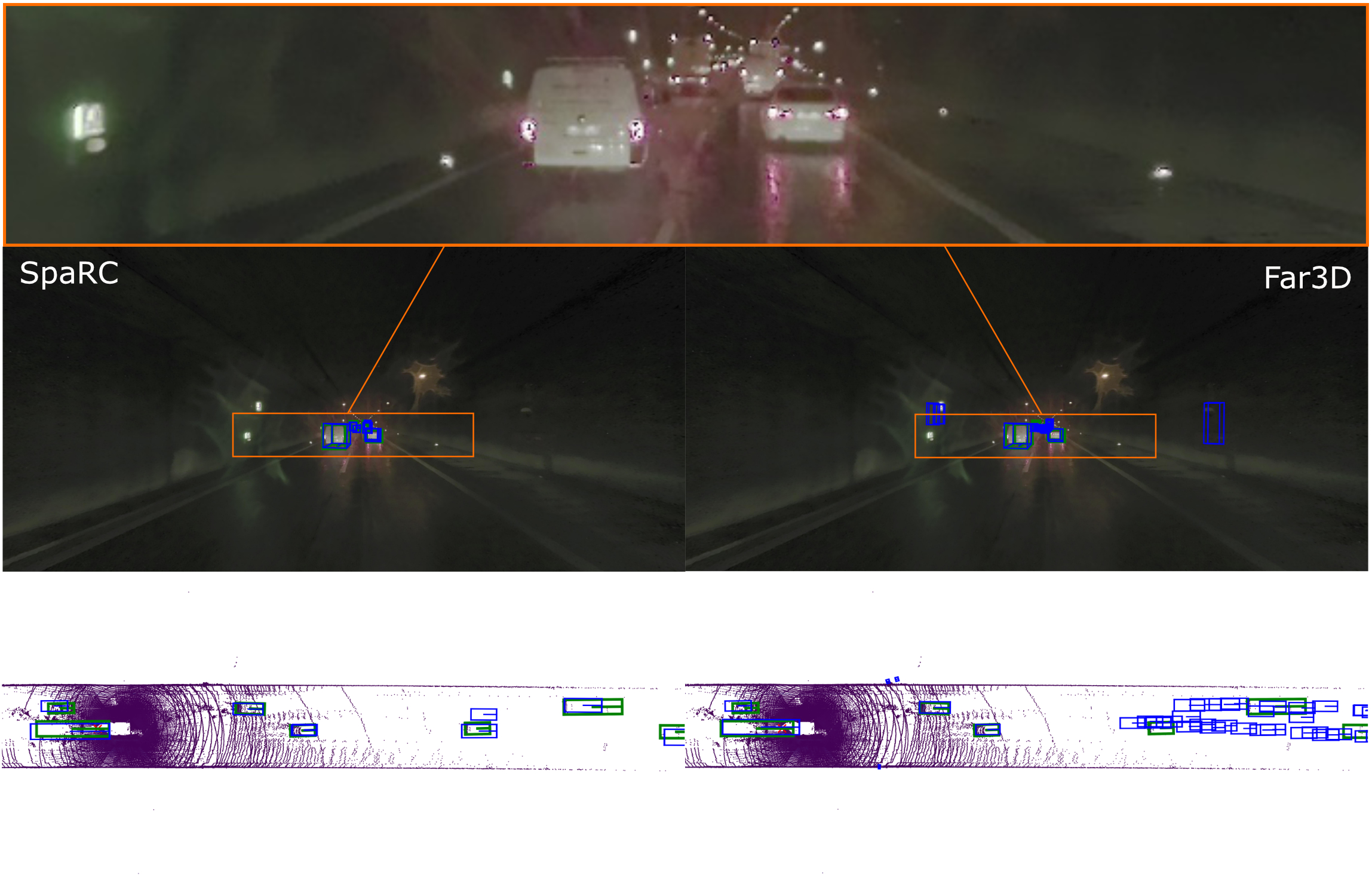}
    \caption{
        \textbf{Qualitative Comparison of SpaRC and Far3D: Dark Tunnel.} 
        We visualize predictions (blue) vs ground truth (green) of 3D object detection on the TruckScenes val set in the perspective and top-down view (150 m detection range). 
        SpaRC is visualized on the left and Far3D on the right. 
        The top row shows a zoomed-in cutout of the long-range detection region for better visibility. 
    }
    \label{fig:comp_2}
\end{figure*}

\begin{figure*}[htbp]
    \centering
    \includegraphics[width=0.82\textwidth]{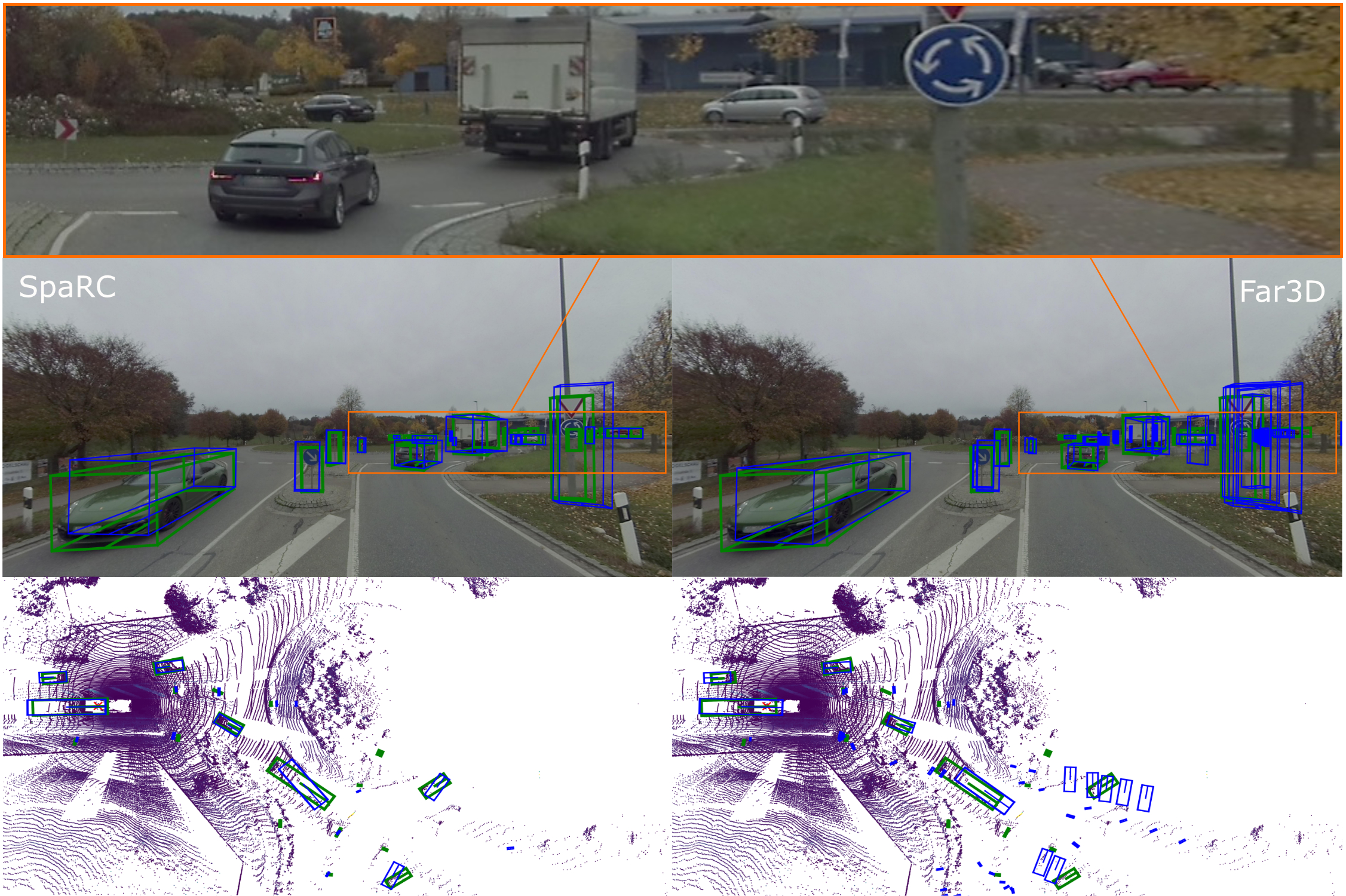}
    \caption{
        \textbf{Qualitative Comparison of SpaRC and Far3D: Roundabout Traffic.} 
        We visualize predictions (blue) vs ground truth (green) of 3D object detection on the TruckScenes val set in the perspective and top-down view (150 m detection range). 
        SpaRC is visualized on the left and Far3D on the right. 
        The top row shows a zoomed-in cutout of the long-range detection region for better visibility. 
    }
    \label{fig:comp_3}
\end{figure*}

\begin{figure*}[htbp]
    \centering
    \includegraphics[width=0.82\textwidth]{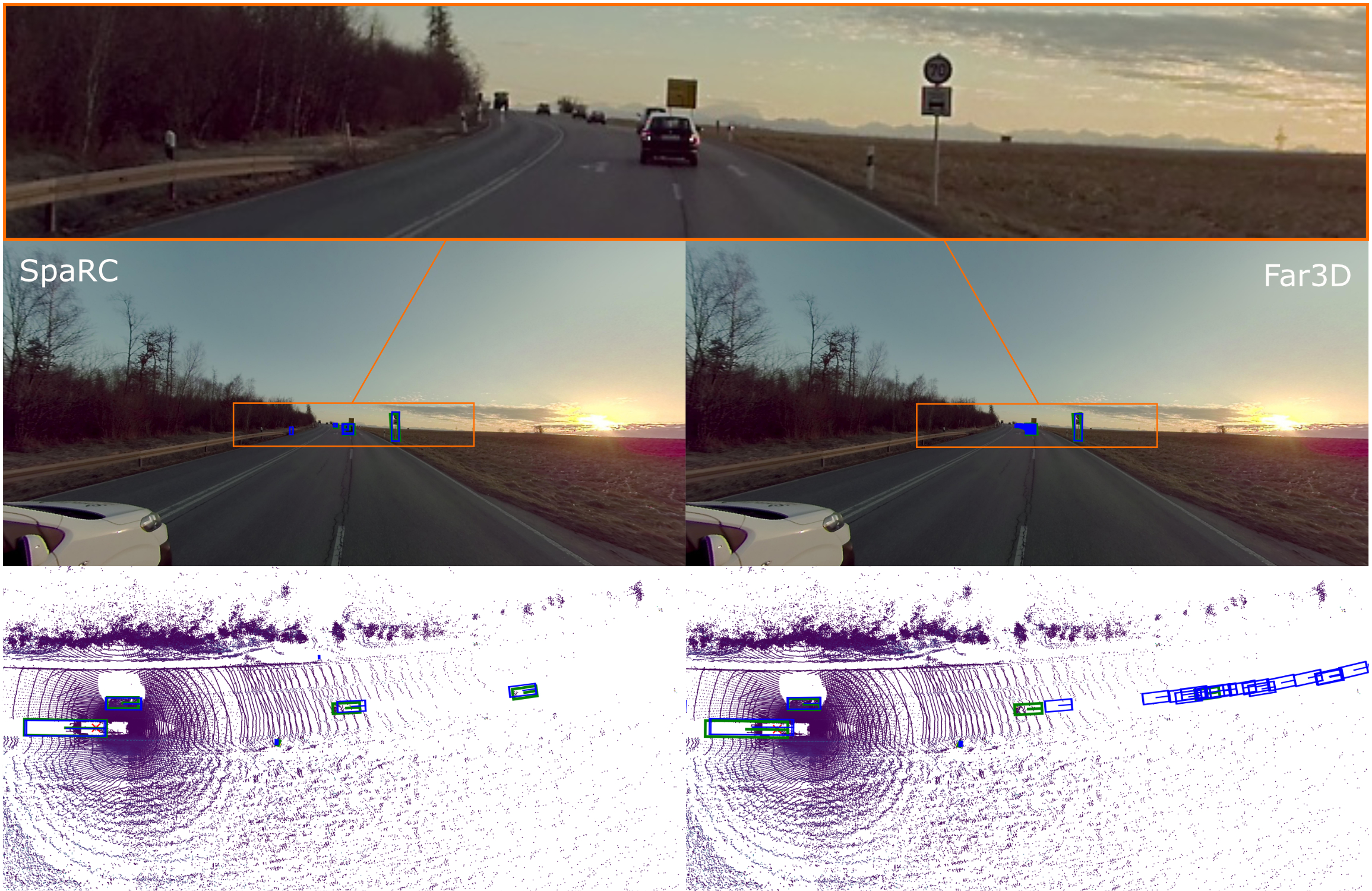}
    \caption{
        \textbf{Qualitative Comparison of SpaRC and Far3D: Sunset Drive.} 
        We visualize predictions (blue) vs ground truth (green) of 3D object detection on the TruckScenes val set in the perspective and top-down view (150 m detection range). 
        SpaRC is visualized on the left and Far3D on the right. 
        The top row shows a zoomed-in cutout of the long-range detection region for better visibility. 
    }
    \label{fig:comp_6}
\end{figure*}

\begin{figure*}[htbp]
    \centering
    \includegraphics[width=0.82\textwidth]{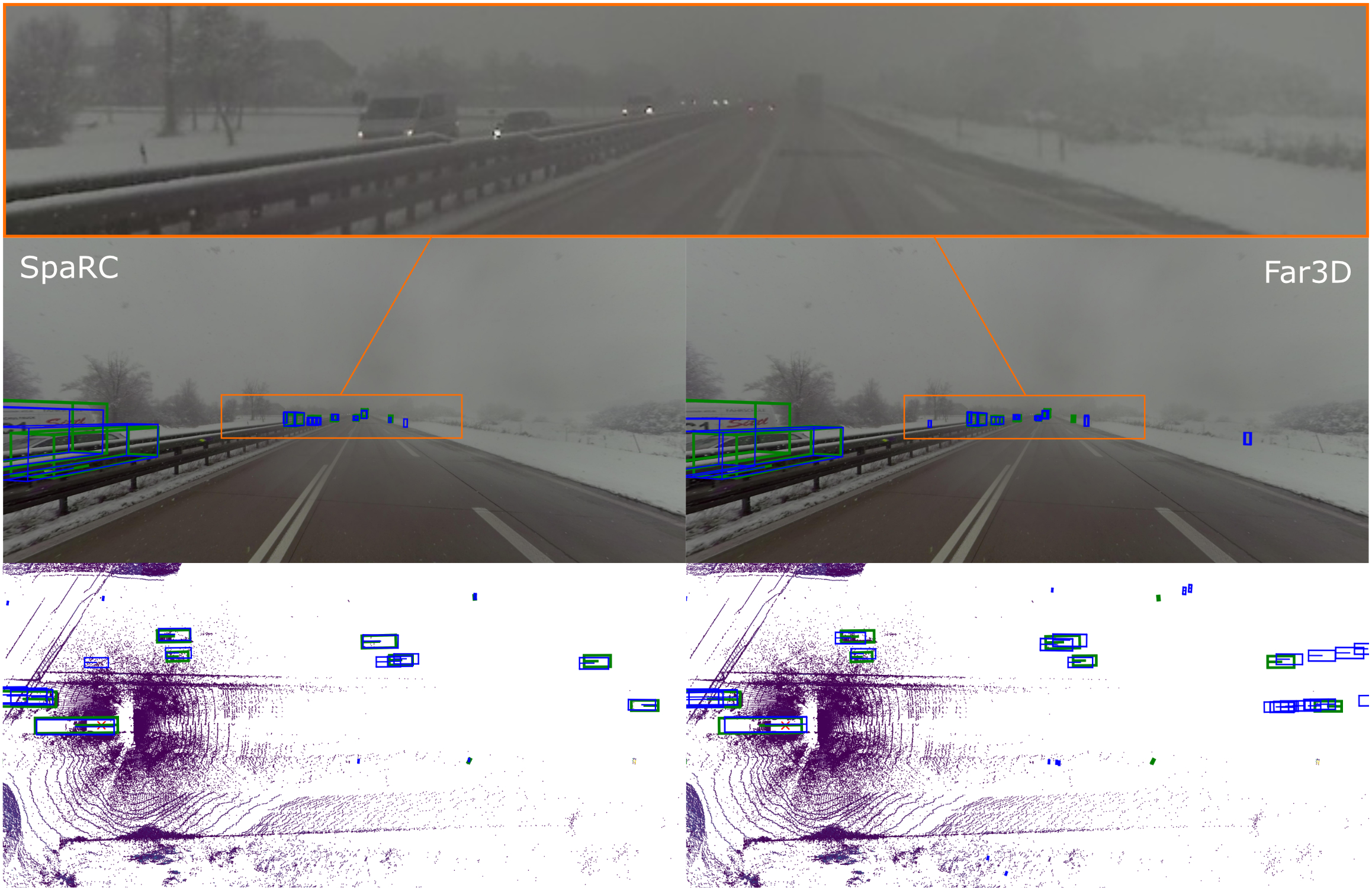}
    \caption{
        \textbf{Qualitative Comparison of SpaRC and Far3D: Winter Scenario.} 
        We visualize predictions (blue) vs ground truth (green) of 3D object detection on the TruckScenes val set in the perspective and top-down view (150 m detection range). 
        SpaRC is visualized on the left and Far3D on the right. 
        The top row shows a zoomed-in cutout of the long-range detection region for better visibility. 
    }
    \label{fig:comp_7}
\end{figure*}

\begin{figure*}[htbp]
    \centering
    \includegraphics[width=0.82\textwidth]{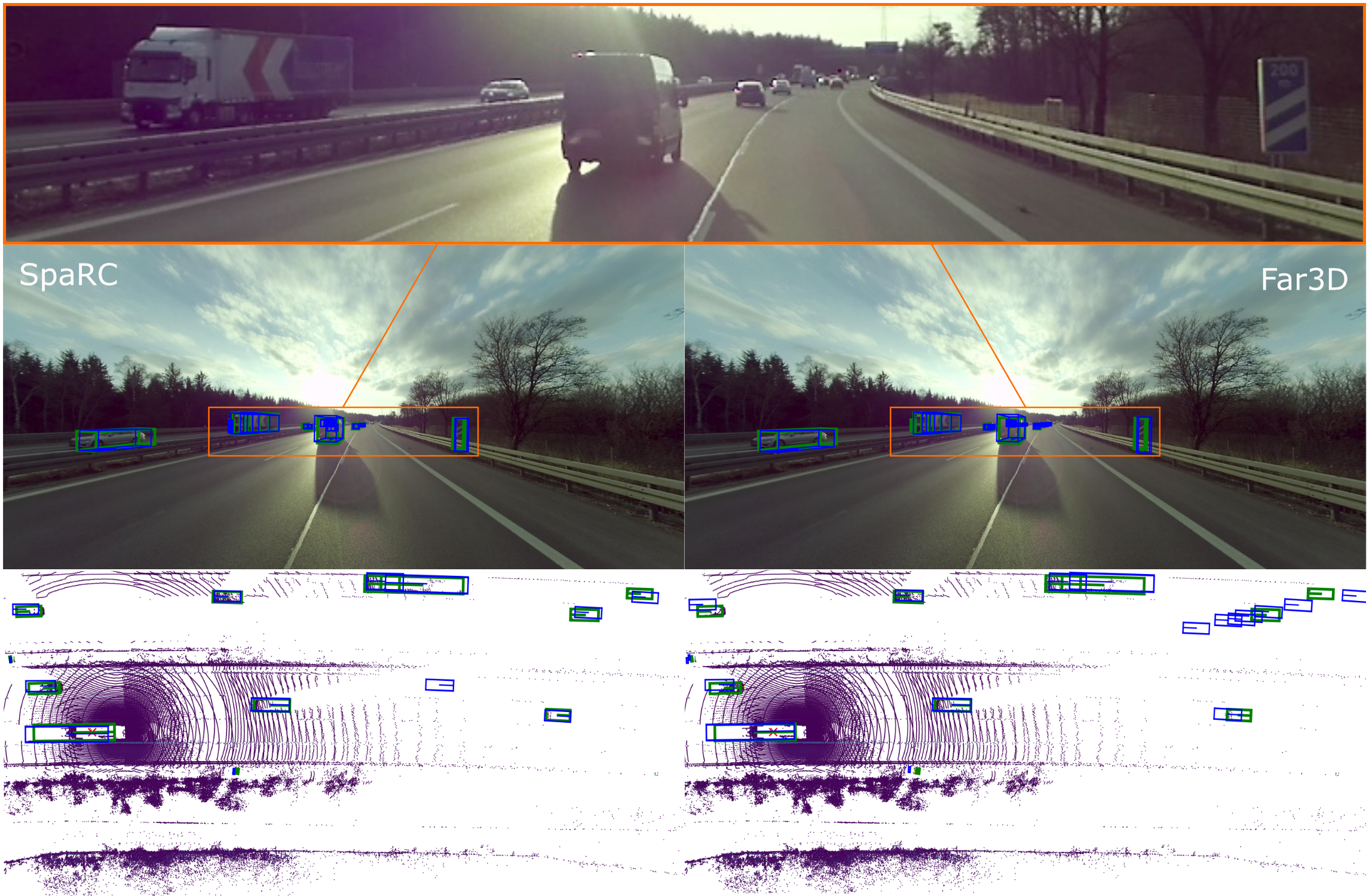}
    \caption{
        \textbf{Qualitative Comparison of SpaRC and Far3D: Glaring Sunlight.} 
        We visualize predictions (blue) vs ground truth (green) of 3D object detection on the TruckScenes val set in the perspective and top-down view (150 m detection range). 
        SpaRC is visualized on the left and Far3D on the right. 
        The top row shows a zoomed-in cutout of the long-range detection region for better visibility. 
    }
    \label{fig:comp_8}
\end{figure*}

The subsequent figures present detailed comparative analyses of 3D object detection performance, comparing predictions from our proposed SpaRC architecture (left panels) with corresponding Far3D outputs (right panels). 
Each figure (\cref{fig:comp_1} - \cref{fig:comp_8}) incorporates both perspective and bird's-eye-view visualizations with a 150-meter detection range. 
We include magnified regions of interest at extended distances in the upper panels to facilitate a detailed examination of long-range detection capabilities.

Our qualitative analysis encompasses a diverse set of challenging environmental conditions, including:
Nighttime scenarios with limited ambient illumination; 
Extended-range detection scenarios beyond 100 meters; 
Adverse weather conditions, including dense fog;
Low-light environments such as tunnels;
Complex traffic scenarios, including roundabout navigation;
Challenging lighting conditions during sunset;
Winter conditions with snow coverage;
High-contrast scenarios with direct solar glare.

%% file: tables/suppl/ptv3_settings.tex
\begin{table}[!t]
    \centering
    \resizebox{0.7\linewidth}{!}{
    \setlength{\tabcolsep}{2mm}
    \begin{tabular}{l|c}
        \toprule
        \textbf{Config} & \textbf{Value} \\
        \midrule
        serialization pattern & Z + TZ + H + TH \\
        patch interaction & Shift Order + Shuffle Order \\
        positional encoding & xCPE \\
        embedding depth & 2 \\
        embedding channels & 32 \\
        encoder depth & [1,1,1,1] \\
        encoder channels & [32, 64, 128, 256] \\
        encoder num heads & [1, 2, 4, 8, 16] \\
        encoder patch size & [64, 64, 64, 64] \\
        decoder depth & [1, 1, 1, 1] \\
        decoder channels & [64, 64, 64, 128] \\
        decoder num heads & [4, 4, 4, 8] \\
        decoder patch size & [64, 64, 64, 64] \\
        down stride & [$\times$2, $\times$2, $\times$2, $\times$2] \\
        mlp ratio & 4 \\
        qkv bias & True \\
        drop path & 0.3 \\
        \bottomrule
    \end{tabular}
    }
    \caption{\textbf{Radar Backbone Settings.}}
    \label{table:ptv3_settings}
\end{table}

%% file: tables/suppl/nusc_inference.tex
\begin{table}[!t]
    \centering
    \resizebox{\linewidth}{!}{
    \setlength{\tabcolsep}{2mm}
    \begin{tabular}{l|c|ccc|cc}
        \toprule
        \textbf{Methods}                            &\textbf{Backbone}      & \textbf{Dec. Lay.}   & \textbf{Queries} & \textbf{R Blocks}    & \textbf{NDS $\uparrow$}         &\textbf{FPS $\uparrow$} \\
        \midrule
        SpaRC Tiny                                 & R18                    & 4                         & 300                 & 2                         & 56.0                 & 32.9 \\
        SpaRC Small                                 & R50                    & 4                         & 300                & 2                         & 60.0                 & 24.3 \\
        SpaRC Base                                 & R50                    & 6                         & 644                 & 3                         & 61.7                 & 19.1 \\
        SpaRC Large                                 & R101                  & 6                         & 644                 & 4                         & 64.4                 & 7.2 \\
        \bottomrule
    \end{tabular}
    }
    \caption{
    \textbf{Model Configurations} for the high-speed inference of SpaRC. We differentiate the model sizes by the different ResNet backbones, the number of decoder layers, the number of queries, and the number of radar-encoding blocks in the point backbone.}
    \label{table:nusc_inference}
    \end{table}

%% file: tables/suppl/trusc_classes.tex
\begin{table*}[!ht]
    \centering
    \resizebox{0.9\linewidth}{!}{
    \setlength{\tabcolsep}{2mm}
    \begin{tabular}{l|c|ccccccccccccc}
        \toprule
        \textbf{Methods}                            & \textbf{Input}        &\textbf{Car}    &\textbf{Truck}       & \textbf{Bus}   & \textbf{Trailer} & \textbf{O.V.} & \textbf{Ped.} & \textbf{M.C.} & \textbf{Bicycle} & \textbf{T.C.} & \textbf{Barrier} & \textbf{Animal} & \textbf{T.S.} \\
        \midrule
        Far3D \cite{jiang2024far3d}                 & C                     & 16.6 & 12.1 & 0.0 & 29.0 & 2.2 & 9.4 & 4.7 & 8.9 & 20.3 & 4.1 & 0.0 & 21.3\\
        \gr \textbf{\sotacolor{SpaRC}}                                    & C+R                   & 43.8 & 29.6 & 1.1 & 40.0 & 8.6 & 18.3 & 16.5 & 14.1 & 40.8 & 14.3 & 0.0 & 41.9\\
        \bottomrule
    \end{tabular}
    }
    \caption{
    \textbf{Per-class Comparison} of 3D object detection (mAP) on TruckScenes val. Abbreviations: O.V. - Other Vehicle, Ped. - Pedestrian, M.C. - Motorcycle, T.C. - Traffic Cone, T.S. - Traffic Sign.}
    \label{table:trusc_classes}
    \end{table*}

%% file: tables/suppl/trusc_range.tex
\begin{table}[!t]
    \centering
    \resizebox{\linewidth}{!}{
    \setlength{\tabcolsep}{2mm}
    \begin{tabular}{l|c | c c c c}
        \toprule
        \textbf{Methods}                            & \textbf{Input}       &\textbf{0 m - 25 m} &\textbf{0 m - 50 m}    &\textbf{0 m - 100 m}       & \textbf{0 m - 150 m} \\
        \midrule
        Far3D \cite{jiang2024far3d}                 & C                     & 30.1 / 34.0               & 23.5 / 28.7              & 14.1 / 23.2             & 10.7 / 21.2\\
        \gr \textbf{\sotacolor{SpaRC}}                                    & C+R                   & 40.9 / 47.0              & 33.8 / 42.9              & 26.9 / 37.7             & 22.5 / 35.4\\
        \bottomrule
    \end{tabular}
    }
    \caption{
    \textbf{Distance Ablation} of \name~on different ranges of the TruckScenes validation set, reporting mAP / NDS.
    }
    \label{table:trusc_range}
    \end{table}

%% file: tables/suppl/nusc_weather.tex
\begin{table}[!t]
    \centering
    \resizebox{\linewidth}{!}{
    \setlength{\tabcolsep}{2mm}
    \begin{tabular}{l|c|c c | c c}
        \toprule
        \textbf{Methods}                            & \textbf{Input}        &\textbf{Sunny}    &\textbf{Rainy}       & \textbf{Day}   & \textbf{Night} \\
        \midrule
        CenterPoint \cite{yin2021center}                & L            &62.9 & 59.2 & 62.8 & 35.4    \\
        \midrule
        RCBEV \cite{zhou2023bridging}                   & C+R                   & 36.1 & 38.5 & 37.1 & 15.5   \\
        BEVDepth \cite{li2022bevdepth}                  & C+R                   & 39.0 & 39.0 & 39.3 & 16.8   \\
        CRN \cite{kim2023crn}                           & C+R                   & 54.8 & 57.0 & 55.1 & 30.4   \\
        \gr \textbf{\sotacolor{SpaRC}}                              & C+R                   & \textbf{57.4} (+2.6) & \textbf{58.6} (+1.6) & \textbf{57.5} (+2.4) & \textbf{36.1} (+5.7) \\
        \bottomrule
    \end{tabular}
    }
    \caption{
    \textbf{Adverse Conditions of nuScenes.} 
    We ablate \name~on the rain and night conditions of the nuScenes validation set. 
    Following \cite{kim2023crn}, we compare the mAP of the R101 configurations.}
    \label{table:nusc_weather}
    \end{table}

%% file: tables/abl_lsa.tex
\begin{table}[!h]
    \centering
    \resizebox{\linewidth}{!}{
    \setlength{\tabcolsep}{4.5mm}
    \begin{tabular}{c|cccccc}
        \toprule
        \textbf{TopK} & \textbf{4} &\textbf{8} &\textbf{16} & \grc\textbf{32} & \textbf{128} & \textbf{644}\\
        \midrule
        mAP & 39.9 & 43.8 & 45.1 & \grc{ \textbf{46.4} } & 46.4 & 45.1 \\
        NDS & 51.5 & 53.7 & 54.2 & \grc{\textbf{54.9}}   & 54.6 & 54.5 \\
        \bottomrule
    \end{tabular}
    }
    \caption{
    \textbf{Ablation of query context} aggregation window in LSA.
    }
    \label{table:lsa}
\end{table}

%% file: tables/suppl/trusc_weather.tex
\setlength{\tabcolsep}{0.65em}
\begin{table*}[ht!]
    \centering
    \resizebox{\linewidth}{!}{%
    \begin{tabular}{l|c|cccc|cccc|cccc|cc}
        \toprule
        \multirow{2}{*}[-0.5ex]{\textbf{Methods}} & \multirow{2}{*}[-0.5ex]{\textbf{mAP}} & \multicolumn{4}{c|}{\textbf{Weather}} & \multicolumn{4}{c|}{\textbf{Daytime}} & \multicolumn{4}{c|}{\textbf{Lighting}}  & \multicolumn{2}{c}{\textbf{Season}}\\
        & & Clear & Rain & Fog & Overcast & Morning & Noon & Evening & Night & Illuminated & Glare & Dark & Twighlight & Summer & Winter \\
        \midrule
        Far3D \cite{jiang2024far3d}                 & 10.7 & 10.6 & 7.0 & 9.6 & 11.2 & 11.2 & 10.4 & 7.4 & 10.6 & 12.0 & 6.7 & 5.7 & 8.9 & 13.2 & 8.4\\
        \gr \textbf{\sotacolor{\name}}                                   & 22.5 & 23.2 & 13.7 & 18.8 & 24.5 & 31.0 & 20.2 & 17.6 & 13.4 & 23.4 & 13.4 & 16.7 & 17.8 & 24.9 & 15.7\\
        \bottomrule
    \end{tabular}%
    }
    \caption{\textbf{Adverse Conditions of TruckScenes}.
        Ablation on different scene tags of TruckScenes validation set. 
    }
    \label{table:trusc_weather}
\end{table*}

%% file: main.bbl
\begin{thebibliography}{83}
\providecommand{\natexlab}[1]{#1}
\providecommand{\url}[1]{\texttt{#1}}
\expandafter\ifx\csname urlstyle\endcsname\relax
  \providecommand{\doi}[1]{doi: #1}\else
  \providecommand{\doi}{doi: \begingroup \urlstyle{rm}\Url}\fi

\bibitem[Armanious et~al.(2024)Armanious, Quach, Ulrich, Winterling, Friesen, Braun, Jenet, Feldman, Kosman, Rapp, et~al.]{armanious2024bosch}
Karim Armanious, Maurice Quach, Michael Ulrich, Timo Winterling, Johannes Friesen, Sascha Braun, Daniel Jenet, Yuri Feldman, Eitan Kosman, Philipp Rapp, et~al.
\newblock Bosch street dataset: A multi-modal dataset with imaging radar for automated driving.
\newblock \emph{arXiv preprint arXiv:2407.12803}, 2024.

\bibitem[Bai et~al.(2022)Bai, Hu, Zhu, Huang, Chen, Fu, and Tai]{bai2022transfusion}
Xuyang Bai, Zeyu Hu, Xinge Zhu, Qingqiu Huang, Yilun Chen, Hongbo Fu, and Chiew-Lan Tai.
\newblock Transfusion: Robust lidar-camera fusion for 3d object detection with transformers.
\newblock In \emph{CVPR}, pages 1090--1099, 2022.

\bibitem[Bang et~al.(2024)Bang, Choi, Kim, Kum, and Choi]{bang2024radardistill}
Geonho Bang, Kwangjin Choi, Jisong Kim, Dongsuk Kum, and Jun~Won Choi.
\newblock Radardistill: Boosting radar-based object detection performance via knowledge distillation from lidar features.
\newblock In \emph{CVPR}, pages 15491--15500, 2024.

\bibitem[Caesar et~al.(2020)Caesar, Bankiti, Lang, Vora, Liong, Xu, Krishnan, Pan, Baldan, and Beijbom]{caesar2020nuscenes}
Holger Caesar, Varun Bankiti, Alex~H Lang, Sourabh Vora, Venice~Erin Liong, Qiang Xu, Anush Krishnan, Yu Pan, Giancarlo Baldan, and Oscar Beijbom.
\newblock nuscenes: A multimodal dataset for autonomous driving.
\newblock In \emph{CVPR}, 2020.

\bibitem[Carion et~al.(2020)Carion, Massa, Synnaeve, Usunier, Kirillov, and Zagoruyko]{carion2020end}
Nicolas Carion, Francisco Massa, Gabriel Synnaeve, Nicolas Usunier, Alexander Kirillov, and Sergey Zagoruyko.
\newblock End-to-end object detection with transformers.
\newblock In \emph{ECCV}, pages 213--229. Springer, 2020.

\bibitem[Chen et~al.(2022)Chen, Zhang, Wang, Wang, and Zhao]{chen2022futr3d}
Xuanyao Chen, Tianyuan Zhang, Yue Wang, Yilun Wang, and Hang Zhao.
\newblock Futr3d: A unified sensor fusion framework for 3d detection.
\newblock In \emph{CVPR}, 2022.

\bibitem[Chen et~al.(2023)Chen, Yu, Chen, Lan, Anandkumar, Jia, and Alvarez]{chen2023focalformer3d}
Yilun Chen, Zhiding Yu, Yukang Chen, Shiyi Lan, Anima Anandkumar, Jiaya Jia, and Jose~M Alvarez.
\newblock Focalformer3d: focusing on hard instance for 3d object detection.
\newblock In \emph{CVPR}, pages 8394--8405, 2023.

\bibitem[Ding et~al.(2024)Ding, Wen, Zhu, Li, and Lu]{ding2024radarocc}
Fangqiang Ding, Xiangyu Wen, Yunzhou Zhu, Yiming Li, and Chris~Xiaoxuan Lu.
\newblock Radarocc: Robust 3d occupancy prediction with 4d imaging radar.
\newblock \emph{arXiv preprint arXiv:2405.14014}, 2024.

\bibitem[Fan et~al.(2024)Fan, Wang, Chang, Li, Wang, and Cao]{fan20244d}
Lili Fan, Junhao Wang, Yuanmeng Chang, Yuke Li, Yutong Wang, and Dongpu Cao.
\newblock 4d mmwave radar for autonomous driving perception: a comprehensive survey.
\newblock \emph{IEEE Transactions on Intelligent Vehicles}, 2024.

\bibitem[Fent et~al.(2023)Fent, Bauerschmidt, and Lienkamp]{fent2023radargnn}
Felix Fent, Philipp Bauerschmidt, and Markus Lienkamp.
\newblock Radargnn: Transformation invariant graph neural network for radar-based perception.
\newblock In \emph{CVPR}, pages 182--191, 2023.

\bibitem[Fent et~al.(2024{\natexlab{a}})Fent, Kuttenreich, Ruch, Rizwin, Juergens, Lechermann, Nissler, Perl, Voll, Yan, et~al.]{fent2024man}
Felix Fent, Fabian Kuttenreich, Florian Ruch, Farija Rizwin, Stefan Juergens, Lorenz Lechermann, Christian Nissler, Andrea Perl, Ulrich Voll, Min Yan, et~al.
\newblock Man truckscenes: A multimodal dataset for autonomous trucking in diverse conditions.
\newblock \emph{NeurIPS}, 2024{\natexlab{a}}.

\bibitem[Fent et~al.(2024{\natexlab{b}})Fent, Palffy, and Caesar]{fent2024dpft}
Felix Fent, Andras Palffy, and Holger Caesar.
\newblock Dpft: Dual perspective fusion transformer for camera-radar-based object detection.
\newblock \emph{arXiv preprint arXiv:2404.03015}, 2024{\natexlab{b}}.

\bibitem[Geiger et~al.(2012)Geiger, Lenz, and Urtasun]{geiger2012we}
Andreas Geiger, Philip Lenz, and Raquel Urtasun.
\newblock Are we ready for autonomous driving? the kitti vision benchmark suite.
\newblock In \emph{CVPR}, pages 3354--3361. IEEE, 2012.

\bibitem[Harley et~al.(2023)Harley, Fang, Li, Ambrus, and Fragkiadaki]{harley2022simple}
Adam~W Harley, Zhaoyuan Fang, Jie Li, Rares Ambrus, and Katerina Fragkiadaki.
\newblock Simple-bev: What really matters for multi-sensor bev perception?
\newblock In \emph{ICRA}, pages 2759--2765, 2023.

\bibitem[He et~al.(2016)He, Zhang, Ren, and Sun]{he2016deep}
Kaiming He, Xiangyu Zhang, Shaoqing Ren, and Jian Sun.
\newblock Deep residual learning for image recognition.
\newblock In \emph{CVPR}, pages 770--778, 2016.

\bibitem[Hou et~al.(2025)Hou, Wang, Ye, Liu, Gong, Tan, Ding, Wang, and Bai]{hou2024open}
Jinghua Hou, Tong Wang, Xiaoqing Ye, Zhe Liu, Shi Gong, Xiao Tan, Errui Ding, Jingdong Wang, and Xiang Bai.
\newblock Open: Object-wise position embedding for multi-view 3d object detection.
\newblock In \emph{ECCV}. Springer, 2025.

\bibitem[Hu et~al.(2023)Hu, Yang, Chen, Li, Sima, Zhu, Chai, Du, Lin, Wang, et~al.]{hu2023planning}
Yihan Hu, Jiazhi Yang, Li Chen, Keyu Li, Chonghao Sima, Xizhou Zhu, Siqi Chai, Senyao Du, Tianwei Lin, Wenhai Wang, et~al.
\newblock Planning-oriented autonomous driving.
\newblock In \emph{CVPR}, pages 17853--17862, 2023.

\bibitem[Huang et~al.(2021)Huang, Huang, Zhu, and Du]{huang2021bevdet}
Junjie Huang, Guan Huang, Zheng Zhu, and Dalong Du.
\newblock Bevdet: High-performance multi-camera 3d object detection in bird-eye-view.
\newblock In \emph{arXiv preprint arXiv:2112.11790}, 2021.

\bibitem[Huang et~al.(2023)Huang, Ye, Liang, Shan, and Du]{huang2023detecting}
Junjie Huang, Yun Ye, Zhujin Liang, Yi Shan, and Dalong Du.
\newblock Detecting as labeling: Rethinking lidar-camera fusion in 3d object detection.
\newblock \emph{arXiv preprint arXiv:2311.07152}, 2023.

\bibitem[Jiang et~al.(2024)Jiang, Li, Liu, Wang, Jia, Wang, Han, and Zhang]{jiang2024far3d}
Xiaohui Jiang, Shuailin Li, Yingfei Liu, Shihao Wang, Fan Jia, Tiancai Wang, Lijin Han, and Xiangyu Zhang.
\newblock Far3d: Expanding the horizon for surround-view 3d object detection.
\newblock In \emph{AAAI}, pages 2561--2569, 2024.

\bibitem[Kim et~al.(2024)Kim, Seong, Bang, Kum, and Choi]{kim2024rcm}
Jisong Kim, Minjae Seong, Geonho Bang, Dongsuk Kum, and Jun~Won Choi.
\newblock Rcm-fusion: Radar-camera multi-level fusion for 3d object detection.
\newblock In \emph{ICRA}, pages 18236--18242. IEEE, 2024.

\bibitem[Kim et~al.(2020)Kim, Choi, and Kum]{kim2020grif}
Youngseok Kim, Jun~Won Choi, and Dongsuk Kum.
\newblock Grif net: Gated region of interest fusion network for robust 3d object detection from radar point cloud and monocular image.
\newblock In \emph{2020 IEEE/RSJ International Conference on Intelligent Robots and Systems (IROS)}, pages 10857--10864. IEEE, 2020.

\bibitem[Kim et~al.(2023{\natexlab{a}})Kim, Kim, Choi, and Kum]{kim2022craft}
Youngseok Kim, Sanmin Kim, Jun~Won Choi, and Dongsuk Kum.
\newblock {CRAFT: Camera-Radar 3D Object Detection with Spatio-Contextual Fusion Transformer}.
\newblock In \emph{AAAI}, 2023{\natexlab{a}}.

\bibitem[Kim et~al.(2023{\natexlab{b}})Kim, Shin, Kim, Lee, Choi, and Kum]{kim2023crn}
Youngseok Kim, Juyeb Shin, Sanmin Kim, In-Jae Lee, Jun~Won Choi, and Dongsuk Kum.
\newblock Crn: Camera radar net for accurate, robust, efficient 3d perception.
\newblock In \emph{ICCV}, 2023{\natexlab{b}}.

\bibitem[Klingner et~al.(2023)Klingner, Borse, Kumar, Rezaei, Narayanan, Yogamani, and Porikli]{klingner2023x3kd}
Marvin Klingner, Shubhankar Borse, Varun~Ravi Kumar, Behnaz Rezaei, Venkatraman Narayanan, Senthil Yogamani, and Fatih Porikli.
\newblock X3kd: Knowledge distillation across modalities, tasks and stages for multi-camera 3d object detection.
\newblock In \emph{CVPR}, 2023.

\bibitem[Lang et~al.(2019)Lang, Vora, Caesar, Zhou, Yang, and Beijbom]{lang2019pointpillars}
Alex~H Lang, Sourabh Vora, Holger Caesar, Lubing Zhou, Jiong Yang, and Oscar Beijbom.
\newblock Pointpillars: Fast encoders for object detection from point clouds.
\newblock In \emph{CVPR}, pages 12697--12705, 2019.

\bibitem[Lee et~al.(2019)Lee, Hwang, Lee, Bae, and Park]{lee2019vov}
Youngwan Lee, Joong-won Hwang, Sangrok Lee, Yuseok Bae, and Jongyoul Park.
\newblock An energy and gpu-computation efficient backbone network for real-time object detection.
\newblock In \emph{CVPRW}, pages 0--0, 2019.

\bibitem[Li et~al.(2023{\natexlab{a}})Li, Sima, Dai, Wang, Lu, Wang, Zeng, Li, Yang, Deng, et~al.]{li2023delving}
Hongyang Li, Chonghao Sima, Jifeng Dai, Wenhai Wang, Lewei Lu, Huijie Wang, Jia Zeng, Zhiqi Li, Jiazhi Yang, Hanming Deng, et~al.
\newblock Delving into the devils of bird's-eye-view perception: A review, evaluation and recipe.
\newblock \emph{IEEE TPAMI}, 2023{\natexlab{a}}.

\bibitem[Li et~al.(2023{\natexlab{b}})Li, Luo, and Yang]{li2023pillarnext}
Jinyu Li, Chenxu Luo, and Xiaodong Yang.
\newblock Pillarnext: Rethinking network designs for 3d object detection in lidar point clouds.
\newblock In \emph{CVPR}, pages 17567--17576, 2023{\natexlab{b}}.

\bibitem[Li et~al.(2022{\natexlab{a}})Li, Chen, Qi, Li, Sun, and Jia]{li2022unifying}
Yanwei Li, Yilun Chen, Xiaojuan Qi, Zeming Li, Jian Sun, and Jiaya Jia.
\newblock Unifying voxel-based representation with transformer for 3d object detection.
\newblock \emph{NeurIPS}, 35, 2022{\natexlab{a}}.

\bibitem[Li et~al.(2023{\natexlab{c}})Li, Bao, Ge, Yang, Sun, and Li]{li2022bevstereo}
Yinhao Li, Han Bao, Zheng Ge, Jinrong Yang, Jianjian Sun, and Zeming Li.
\newblock Bevstereo: Enhancing depth estimation in multi-view 3d object detection with dynamic temporal stereo.
\newblock In \emph{AAAI}, 2023{\natexlab{c}}.

\bibitem[Li et~al.(2023{\natexlab{d}})Li, Ge, Yu, Yang, Wang, Shi, Sun, and Li]{li2022bevdepth}
Yinhao Li, Zheng Ge, Guanyi Yu, Jinrong Yang, Zengran Wang, Yukang Shi, Jianjian Sun, and Zeming Li.
\newblock Bevdepth: Acquisition of reliable depth for multi-view 3d object detection.
\newblock In \emph{AAAI}, 2023{\natexlab{d}}.

\bibitem[Li et~al.(2022{\natexlab{b}})Li, Wang, Li, Xie, Sima, Lu, Yu, and Dai]{li2022bevformer}
Zhiqi Li, Wenhai Wang, Hongyang Li, Enze Xie, Chonghao Sima, Tong Lu, Qiao Yu, and Jifeng Dai.
\newblock Bevformer: Learning bird's-eye-view representation from multi-camera images via spatiotemporal transformers.
\newblock In \emph{ECCV}, 2022{\natexlab{b}}.

\bibitem[Li et~al.(2023{\natexlab{e}})Li, Yu, Austin, Fang, Lan, Kautz, and Alvarez]{li2023fbocc}
Zhiqi Li, Zhiding Yu, David Austin, Mingsheng Fang, Shiyi Lan, Jan Kautz, and Jose~M Alvarez.
\newblock Fb-occ: 3d occupancy prediction based on forward-backward view transformation.
\newblock \emph{arXiv preprint arXiv:2307.01492}, 2023{\natexlab{e}}.

\bibitem[Li et~al.(2023{\natexlab{f}})Li, Yu, Wang, Anandkumar, Lu, and Alvarez]{li2023fb}
Zhiqi Li, Zhiding Yu, Wenhai Wang, Anima Anandkumar, Tong Lu, and Jose~M Alvarez.
\newblock Fb-bev: Bev representation from forward-backward view transformations.
\newblock In \emph{ICCV}, 2023{\natexlab{f}}.

\bibitem[Li et~al.(2024)Li, Lan, Alvarez, and Wu]{li2024bevnext}
Zhenxin Li, Shiyi Lan, Jose~M Alvarez, and Zuxuan Wu.
\newblock Bevnext: Reviving dense bev frameworks for 3d object detection.
\newblock In \emph{CVPR}, pages 20113--20123, 2024.

\bibitem[Liang et~al.(2022)Liang, Xie, Yu, Xia, Lin, Wang, Tang, Wang, and Tang]{liang2022bevfusion}
Tingting Liang, Hongwei Xie, Kaicheng Yu, Zhongyu Xia, Zhiwei Lin, Yongtao Wang, Tao Tang, Bing Wang, and Zhi Tang.
\newblock Bevfusion: A simple and robust lidar-camera fusion framework.
\newblock \emph{NeurIPS}, 35, 2022.

\bibitem[Lin et~al.(2017)Lin, Doll{\'a}r, Girshick, He, Hariharan, and Belongie]{lin2017feature}
Tsung-Yi Lin, Piotr Doll{\'a}r, Ross Girshick, Kaiming He, Bharath Hariharan, and Serge Belongie.
\newblock Feature pyramid networks for object detection.
\newblock In \emph{CVPR}, pages 2117--2125, 2017.

\bibitem[Lin et~al.(2022)Lin, Lin, Pei, Huang, and Su]{lin2022sparse4d}
Xuewu Lin, Tianwei Lin, Zixiang Pei, Lichao Huang, and Zhizhong Su.
\newblock Sparse4d: Multi-view 3d object detection with sparse spatial-temporal fusion.
\newblock \emph{arXiv preprint arXiv:2211.10581}, 2022.

\bibitem[Lin et~al.(2024)Lin, Liu, Xia, Wang, Wang, Qi, Dong, Dong, Zhang, and Zhu]{lin2024rcbevdet}
Zhiwei Lin, Zhe Liu, Zhongyu Xia, Xinhao Wang, Yongtao Wang, Shengxiang Qi, Yang Dong, Nan Dong, Le Zhang, and Ce Zhu.
\newblock Rcbevdet: Radar-camera fusion in bird's eye view for 3d object detection.
\newblock In \emph{CVPR}, pages 14928--14937, 2024.

\bibitem[Liu et~al.(2025)Liu, Huang, Zhang, Yao, Zhang, Wan, Ye, and Zhou]{liu2025ray}
Feng Liu, Tengteng Huang, Qianjing Zhang, Haotian Yao, Chi Zhang, Fang Wan, Qixiang Ye, and Yanzhao Zhou.
\newblock Ray denoising: Depth-aware hard negative sampling for multi-view 3d object detection.
\newblock In \emph{ECCV}, pages 200--217. Springer, 2025.

\bibitem[Liu et~al.(2023{\natexlab{a}})Liu, Teng, Lu, Wang, and Wang]{liu2023sparsebev}
Haisong Liu, Yao Teng, Tao Lu, Haiguang Wang, and Limin Wang.
\newblock Sparsebev: High-performance sparse 3d object detection from multi-camera videos.
\newblock In \emph{ICCV}, 2023{\natexlab{a}}.

\bibitem[Liu et~al.(2022)Liu, Wang, Zhang, and Sun]{liu2022petr}
Yingfei Liu, Tiancai Wang, Xiangyu Zhang, and Jian Sun.
\newblock Petr: Position embedding transformation for multi-view 3d object detection.
\newblock In \emph{ECCV}, pages 531--–548, 2022.

\bibitem[Liu et~al.(2023{\natexlab{b}})Liu, Yan, Jia, Li, Gao, Wang, and Zhang]{liu2023petrv2}
Yingfei Liu, Junjie Yan, Fan Jia, Shuailin Li, Aqi Gao, Tiancai Wang, and Xiangyu Zhang.
\newblock Petrv2: A unified framework for 3d perception from multi-camera images.
\newblock In \emph{ICCV}, pages 3262--3272, 2023{\natexlab{b}}.

\bibitem[Liu et~al.(2023{\natexlab{c}})Liu, Tang, Amini, Yang, Mao, Rus, and Han]{liu2022bevfusion}
Zhijian Liu, Haotian Tang, Alexander Amini, Xinyu Yang, Huizi Mao, Daniela Rus, and Song Han.
\newblock Bevfusion: Multi-task multi-sensor fusion with unified bird's-eye view representation.
\newblock In \emph{ICRA}, 2023{\natexlab{c}}.

\bibitem[Long et~al.(2023)Long, Kumar, Morris, Liu, Castro, and Chakravarty]{long2023radiant}
Yunfei Long, Abhinav Kumar, Daniel Morris, Xiaoming Liu, Marcos Castro, and Punarjay Chakravarty.
\newblock Radiant: Radar-image association network for 3d object detection.
\newblock In \emph{AAAI}, 2023.

\bibitem[Loshchilov and Hutter(2017)]{loshchilov2017decoupled}
Ilya Loshchilov and Frank Hutter.
\newblock Decoupled weight decay regularization.
\newblock \emph{arXiv preprint arXiv:1711.05101}, 2017.

\bibitem[Man et~al.(2023)Man, Gui, and Wang]{man2023bev}
Yunze Man, Liang-Yan Gui, and Yu-Xiong Wang.
\newblock Bev-guided multi-modality fusion for driving perception.
\newblock In \emph{CVPR}, 2023.

\bibitem[Musiat et~al.(2024)Musiat, Reichardt, Schulze, and Wasenm{\"u}ller]{musiat2024radarpillars}
Alexander Musiat, Laurenz Reichardt, Michael Schulze, and Oliver Wasenm{\"u}ller.
\newblock Radarpillars: Efficient object detection from 4d radar point clouds.
\newblock \emph{arXiv preprint arXiv:2408.05020}, 2024.

\bibitem[Nabati and Qi(2021)]{nabati2021centerfusion}
Ramin Nabati and Hairong Qi.
\newblock Centerfusion: Center-based radar and camera fusion for 3d object detection.
\newblock In \emph{WACV}, pages 1527--1536, 2021.

\bibitem[Nobis et~al.(2019)Nobis, Geisslinger, Weber, Betz, and Lienkamp]{nobis19crfnet}
Felix Nobis, Maximilian Geisslinger, Markus Weber, Johannes Betz, and Markus Lienkamp.
\newblock A deep learning-based radar and camera sensor fusion architecture for object detection.
\newblock In \emph{2019 Sensor Data Fusion: Trends, Solutions, Applications (SDF)}, 2019.

\bibitem[Paek et~al.(2022)Paek, Kong, and Wijaya]{paek2022k}
Dong-Hee Paek, Seung-Hyun Kong, and Kevin~Tirta Wijaya.
\newblock K-radar: 4d radar object detection for autonomous driving in various weather conditions.
\newblock \emph{NeurIPS}, 35:\penalty0 3819--3829, 2022.

\bibitem[Park et~al.(2023)Park, Xu, Yang, Keutzer, Kitani, Tomizuka, and Zhan]{park2022time}
Jinhyung Park, Chenfeng Xu, Shijia Yang, Kurt Keutzer, Kris Kitani, Masayoshi Tomizuka, and Wei Zhan.
\newblock Time will tell: New outlooks and a baseline for temporal multi-view 3d object detection.
\newblock In \emph{ICLR}, 2023.

\bibitem[Philion and Fidler(2020)]{philion2020lift}
Jonah Philion and Sanja Fidler.
\newblock Lift, splat, shoot: Encoding images from arbitrary camera rigs by implicitly unprojecting to 3d.
\newblock In \emph{ECCV}, pages 194--210, 2020.

\bibitem[Sun et~al.(2020)Sun, Kretzschmar, Dotiwalla, Chouard, Patnaik, Tsui, Guo, Zhou, Chai, Caine, et~al.]{sun2020waymo}
Pei Sun, Henrik Kretzschmar, Xerxes Dotiwalla, Aurelien Chouard, Vijaysai Patnaik, Paul Tsui, James Guo, Yin Zhou, Yuning Chai, Benjamin Caine, et~al.
\newblock Scalability in perception for autonomous driving: Waymo open dataset.
\newblock In \emph{CVPR}, pages 2446--2454, 2020.

\bibitem[Tang et~al.(2025)Tang, Meng, Chen, and Cheng]{tang2025simpb}
Yingqi Tang, Zhaotie Meng, Guoliang Chen, and Erkang Cheng.
\newblock Simpb: A single model for 2d and 3d object detection from multiple cameras.
\newblock In \emph{ECCV}. Springer, 2025.

\bibitem[Teepe et~al.(2024{\natexlab{a}})Teepe, Wolters, Gilg, Herzog, and Rigoll]{teepe2024earlybird}
Torben Teepe, Philipp Wolters, Johannes Gilg, Fabian Herzog, and Gerhard Rigoll.
\newblock Earlybird: Early-fusion for multi-view tracking in the bird's eye view.
\newblock In \emph{WACV}, pages 102--111, 2024{\natexlab{a}}.

\bibitem[Teepe et~al.(2024{\natexlab{b}})Teepe, Wolters, Gilg, Herzog, and Rigoll]{teepe2024lifting}
Torben Teepe, Philipp Wolters, Johannes Gilg, Fabian Herzog, and Gerhard Rigoll.
\newblock Lifting multi-view detection and tracking to the bird's eye view.
\newblock In \emph{CVPRW}, pages 667--676, 2024{\natexlab{b}}.

\bibitem[Ulrich et~al.(2022)Ulrich, Braun, K{\"o}hler, Niederl{\"o}hner, Faion, Gl{\"a}ser, and Blume]{ulrich2022impr}
Michael Ulrich, Sascha Braun, Daniel K{\"o}hler, Daniel Niederl{\"o}hner, Florian Faion, Claudius Gl{\"a}ser, and Holger Blume.
\newblock Improved orientation estimation and detection with hybrid object detection networks for automotive radar.
\newblock In \emph{Proceedings of the IEEE International Intelligent Transportation Systems Conference (ITSC)}, 2022.

\bibitem[Wang et~al.(2023{\natexlab{a}})Wang, Shi, Shi, Lei, Wang, He, Schiele, and Wang]{wang2023dsvt}
Haiyang Wang, Chen Shi, Shaoshuai Shi, Meng Lei, Sen Wang, Di He, Bernt Schiele, and Liwei Wang.
\newblock Dsvt: Dynamic sparse voxel transformer with rotated sets.
\newblock In \emph{CVPR}, pages 13520--13529, 2023{\natexlab{a}}.

\bibitem[Wang et~al.(2023{\natexlab{b}})Wang, Liu, Wang, Li, and Zhang]{Wang2023streampetr}
Shihao Wang, Yingfei Liu, Tiancai Wang, Ying Li, and Xiangyu Zhang.
\newblock Exploring object-centric temporal modeling for efficient multi-view 3d object detection.
\newblock In \emph{ICCV}, 2023{\natexlab{b}}.

\bibitem[Wang et~al.(2021)Wang, Zhu, Pang, and Lin]{wang2021fcos3d}
Tai Wang, Xinge Zhu, Jiangmiao Pang, and Dahua Lin.
\newblock Fcos3d: Fully convolutional one-stage monocular 3d object detection.
\newblock In \emph{ICCV}, pages 913--922, 2021.

\bibitem[Wang et~al.(2022{\natexlab{a}})Wang, Xinge, Pang, and Lin]{wang2022probabilistic}
Tai Wang, ZHU Xinge, Jiangmiao Pang, and Dahua Lin.
\newblock Probabilistic and geometric depth: Detecting objects in perspective.
\newblock In \emph{Conference on Robot Learning}, pages 1475--1485. PMLR, 2022{\natexlab{a}}.

\bibitem[Wang et~al.(2022{\natexlab{b}})Wang, Guizilini, Zhang, Wang, Zhao, and Solomon]{wang2022detr3d}
Yue Wang, Vitor~Campagnolo Guizilini, Tianyuan Zhang, Yilun Wang, Hang Zhao, and Justin Solomon.
\newblock Detr3d: 3d object detection from multi-view images via 3d-to-2d queries.
\newblock In \emph{Conference on Robot Learning}, pages 180--191. PMLR, 2022{\natexlab{b}}.

\bibitem[Weng and Kitani(2019)]{weng2019baseline}
Xinshuo Weng and Kris Kitani.
\newblock A baseline for 3d multi-object tracking.
\newblock \emph{arXiv preprint arXiv:1907.03961}, 1\penalty0 (2):\penalty0 6, 2019.

\bibitem[Wilson et~al.(2023)Wilson, Qi, Agarwal, Lambert, Singh, Khandelwal, Pan, Kumar, Hartnett, Pontes, et~al.]{wilson2023argoverse}
Benjamin Wilson, William Qi, Tanmay Agarwal, John Lambert, Jagjeet Singh, Siddhesh Khandelwal, Bowen Pan, Ratnesh Kumar, Andrew Hartnett, Jhony~Kaesemodel Pontes, et~al.
\newblock Argoverse 2: Next generation datasets for self-driving perception and forecasting.
\newblock \emph{arXiv preprint arXiv:2301.00493}, 2023.

\bibitem[Wolters et~al.(2025)Wolters, Gilg, Teepe, Herzog, Laouichi, Hofmann, and Rigoll]{wolters2024hydra}
Philipp Wolters, Johannes Gilg, Torben Teepe, Fabian Herzog, Anouar Laouichi, Martin Hofmann, and Gerhard Rigoll.
\newblock Unleashing hydra: Hybrid fusion, depth consistency and radar for unified 3d perception.
\newblock In \emph{2025 IEEE International Conference on Robotics and Automation (ICRA)}, pages 7467--7474. IEEE, 2025.

\bibitem[Wu et~al.(2024)Wu, Jiang, Wang, Liu, Liu, Qiao, Ouyang, He, and Zhao]{wu2024point}
Xiaoyang Wu, Li Jiang, Peng-Shuai Wang, Zhijian Liu, Xihui Liu, Yu Qiao, Wanli Ouyang, Tong He, and Hengshuang Zhao.
\newblock Point transformer v3: Simpler faster stronger.
\newblock In \emph{CVPR}, pages 4840--4851, 2024.

\bibitem[Wu et~al.(2023)Wu, Chen, Gan, Wang, and Pu]{wu2023mvfusion}
Zizhang Wu, Guilian Chen, Yuanzhu Gan, Lei Wang, and Jian Pu.
\newblock Mvfusion: Multi-view 3d object detection with semantic-aligned radar and camera fusion.
\newblock In \emph{ICRA}, 2023.

\bibitem[Xu et~al.(2023)Xu, Kong, Shuai, and Liu]{xu2023frnet}
Xiang Xu, Lingdong Kong, Hui Shuai, and Qingshan Liu.
\newblock Frnet: Frustum-range networks for scalable lidar segmentation.
\newblock \emph{arXiv preprint arXiv:2312.04484}, 2023.

\bibitem[Yan et~al.(2023)Yan, Liu, Sun, Jia, Li, Wang, and Zhang]{yan2023cmt}
Junjie Yan, Yingfei Liu, Jianjian Sun, Fan Jia, Shuailin Li, Tiancai Wang, and Xiangyu Zhang.
\newblock Cross modal transformer: Towards fast and robust 3d object detection.
\newblock In \emph{ICCV}, pages 18268--18278, 2023.

\bibitem[Yang et~al.(2023)Yang, Chen, Tian, Tao, Zhu, Zhang, Huang, Li, Qiao, Lu, et~al.]{yang2023bevformerV2}
Chenyu Yang, Yuntao Chen, Hao Tian, Chenxin Tao, Xizhou Zhu, Zhaoxiang Zhang, Gao Huang, Hongyang Li, Yu Qiao, Lewei Lu, et~al.
\newblock Bevformer v2: Adapting modern image backbones to bird's-eye-view recognition via perspective supervision.
\newblock In \emph{CVPR}, 2023.

\bibitem[Yang et~al.(2022{\natexlab{a}})Yang, Yu, Li, Li, and Tao]{yang2022quality}
Jinrong Yang, En Yu, Zeming Li, Xiaoping Li, and Wenbing Tao.
\newblock Quality matters: Embracing quality clues for robust 3d multi-object tracking.
\newblock \emph{arXiv preprint arXiv:2208.10976}, 2022{\natexlab{a}}.

\bibitem[Yang et~al.(2022{\natexlab{b}})Yang, Chen, Miao, Li, Zhu, and Zhang]{yang2022deepinteraction}
Zeyu Yang, Jiaqi Chen, Zhenwei Miao, Wei Li, Xiatian Zhu, and Li Zhang.
\newblock Deepinteraction: 3d object detection via modality interaction.
\newblock \emph{NeurIPS}, 35:\penalty0 1992--2005, 2022{\natexlab{b}}.

\bibitem[Yin et~al.(2021)Yin, Zhou, and Krahenbuhl]{yin2021center}
Tianwei Yin, Xingyi Zhou, and Philipp Krahenbuhl.
\newblock Center-based 3d object detection and tracking.
\newblock In \emph{CVPR}, pages 11784--11793, 2021.

\bibitem[Yoneda et~al.(2019)Yoneda, Suganuma, Yanase, and Aldibaja]{yoneda2019automated}
Keisuke Yoneda, Naoki Suganuma, Ryo Yanase, and Mohammad Aldibaja.
\newblock Automated driving recognition technologies for adverse weather conditions.
\newblock \emph{IATSS research}, 43\penalty0 (4):\penalty0 253--262, 2019.

\bibitem[Zhang et~al.(2023{\natexlab{a}})Zhang, Li, Liu, Zhang, Su, Zhu, Ni, and Shum]{zhang2023dino}
Hao Zhang, Feng Li, Shilong Liu, Lei Zhang, Hang Su, Jun Zhu, Lionel Ni, and Heung-Yeung Shum.
\newblock {DINO}: {DETR} with improved denoising anchor boxes for end-to-end object detection.
\newblock In \emph{ICLR}, 2023{\natexlab{a}}.

\bibitem[Zhang et~al.(2023{\natexlab{b}})Zhang, Wang, Ye, Zhang, Lu, Tan, Ding, Sun, and Wang]{zhang2023bytetrackv2}
Yifu Zhang, Xinggang Wang, Xiaoqing Ye, Wei Zhang, Jincheng Lu, Xiao Tan, Errui Ding, Peize Sun, and Jingdong Wang.
\newblock Bytetrackv2: 2d and 3d multi-object tracking by associating every detection box.
\newblock \emph{arXiv preprint arXiv:2303.15334}, 2023{\natexlab{b}}.

\bibitem[Zhao et~al.(2021)Zhao, Jiang, Jia, Torr, and Koltun]{zhao2021point}
Hengshuang Zhao, Li Jiang, Jiaya Jia, Philip~HS Torr, and Vladlen Koltun.
\newblock Point transformer.
\newblock In \emph{ICCV}, pages 16259--16268, 2021.

\bibitem[Zhao et~al.(2024)Zhao, Song, and Skinner]{zhao2024crkd}
Lingjun Zhao, Jingyu Song, and Katherine~A Skinner.
\newblock Crkd: Enhanced camera-radar object detection with cross-modality knowledge distillation.
\newblock In \emph{CVPR}, 2024.

\bibitem[Zhou et~al.(2023)Zhou, Chen, Shi, Jiang, Yang, and Yang]{zhou2023bridging}
Taohua Zhou, Junjie Chen, Yining Shi, Kun Jiang, Mengmeng Yang, and Diange Yang.
\newblock Bridging the view disparity between radar and camera features for multi-modal fusion 3d object detection.
\newblock \emph{IEEE Transactions on Intelligent Vehicles}, 2023.

\bibitem[Zhou et~al.(2019)Zhou, Wang, and Kr{\"a}henb{\"u}hl]{zhou2019objects}
Xingyi Zhou, Dequan Wang, and Philipp Kr{\"a}henb{\"u}hl.
\newblock Objects as points.
\newblock \emph{arXiv preprint arXiv:1904.07850}, 2019.

\bibitem[Zhu et~al.(2019)Zhu, Jiang, Zhou, Li, and Yu]{zhu2019class}
Benjin Zhu, Zhengkai Jiang, Xiangxin Zhou, Zeming Li, and Gang Yu.
\newblock Class-balanced grouping and sampling for point cloud 3d object detection.
\newblock \emph{arXiv preprint arXiv:1908.09492}, 2019.

\end{thebibliography}
